\documentclass[10pt,twocolumn,letterpaper]{article}

\usepackage[pagenumbers]{cvpr} 

\definecolor{cvprblue}{rgb}{0.21,0.49,0.74}
\usepackage[pagebackref,breaklinks,colorlinks,allcolors=cvprblue]{hyperref}

\makeatletter
\DeclareRobustCommand\onedot{\futurelet\@let@token\@onedot}
\def\@onedot{\ifx\@let@token.\else.\null\fi\xspace}

\usepackage{multirow}
\usepackage{multicol}
\usepackage{tipa}
\usepackage[accsupp]{axessibility}  

\newcommand{\myParagraph}[1]{\noindent \textbf{#1}}

\def\paperID{8178} 
\def\confName{CVPR}
\def\confYear{2025}

\title{
Reloc3r: Large-Scale Training of Relative Camera Pose Regression for Generalizable, Fast, and Accurate Visual Localization 
}

\author{
Siyan Dong$^{1}$\thanks{Joint first authors: siyan3d@hku.hk, shuzhe.wang@aalto.fi} \quad
Shuzhe Wang$^{2}$\footnotemark[1] \quad
Shaohui Liu$^{3}$ \quad
Lulu Cai$^{1}$ \\
Qingnan Fan$^{4}$ \quad
Juho Kannala$^{2,5}$\thanks{Corresponding authors} \quad
Yanchao Yang$^{1}$\footnotemark[2] \\
$^1${The University of Hong Kong} \quad
$^2${Aalto University} \quad 
$^3${ETH Zurich} \quad 
$^4${VIVO} \quad
$^5${University of Oulu} 
}

\begin{document}
\maketitle
\begin{abstract}

Visual localization aims to determine the camera pose of a query image relative to a database of posed images. In recent years, deep neural networks that directly regress camera poses have gained popularity due to their fast inference capabilities. However, existing methods struggle to either generalize well to new scenes or provide accurate camera pose estimates. To address these issues, we present \textbf{Reloc3r}, a simple yet effective visual localization framework. It consists of an elegantly designed relative pose regression network, and a minimalist motion averaging module for absolute pose estimation. Trained on approximately eight million posed image pairs, Reloc3r achieves surprisingly good performance and generalization ability. We conduct extensive experiments on six public datasets, consistently demonstrating the effectiveness and efficiency of the proposed method. It provides high-quality camera pose estimates in real time and generalizes to novel scenes. Code: \url{https://github.com/ffrivera0/reloc3r}.

\end{abstract}

\section{Introduction}
\label{sec:intro}

\begin{figure}[!t]
\centering
\includegraphics[width=0.99\linewidth]{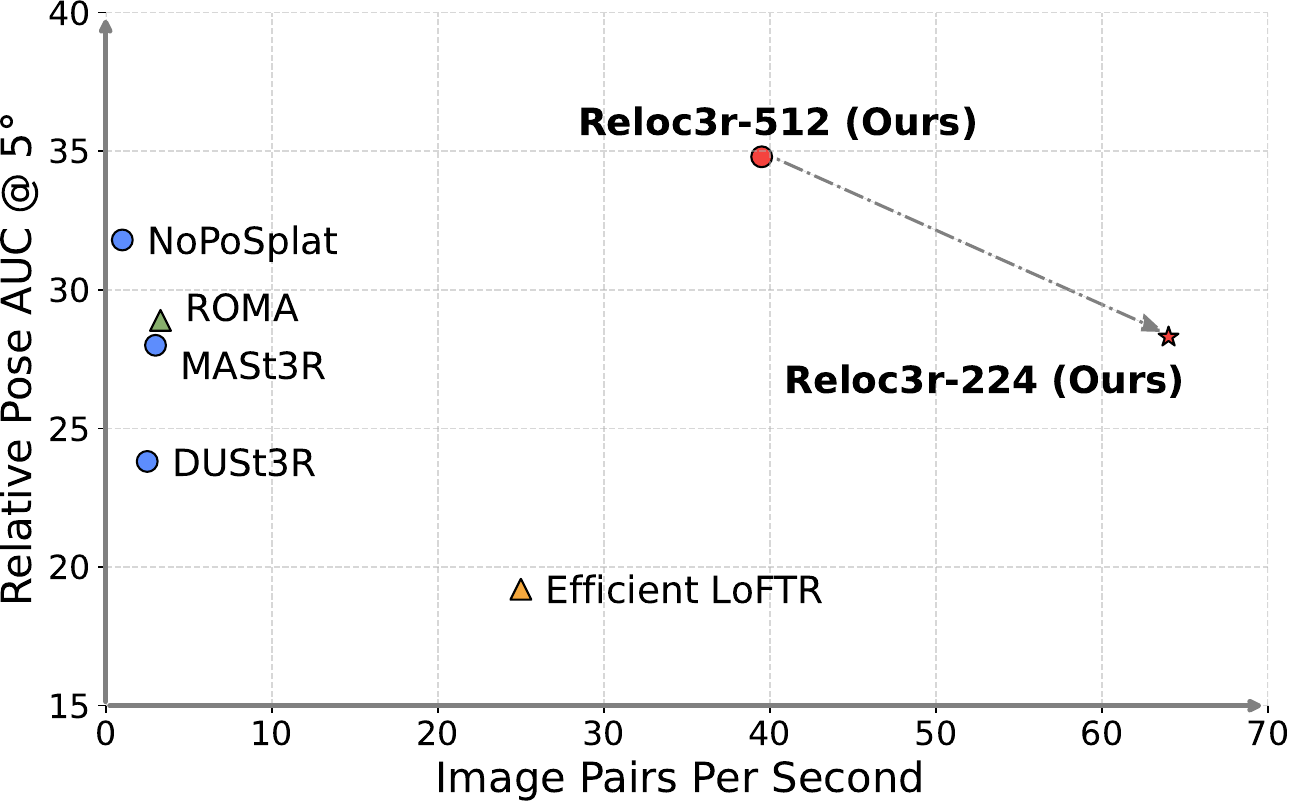}
\caption{
Comparison of pose accuracy and runtime efficiency.
We report the AUC@5 and image pairs per second (FPS) on the ScanNet1500~\cite{dai2017scannet,sarlin2020superglue} dataset. 
We provide two versions of Reloc3r: one trained and tested on image widths of 512, and another on 224. The proposed Reloc3r-512 outperforms all other methods, achieving the best AUC@5 while maintaining an efficiency of 24 FPS. Remarkably, even at 224 resolution, our method matches ROMA~\cite{edstedt2024roma} in accuracy while being 20$\times$ faster.
}
\label{fig:teaser}
\end{figure}

Visual localization, also known as camera re-localization, is a key challenge in computer vision, robotics, and graphics. It's crucial for many applications, including augmented reality and robot navigation. The process involves registering new query images to a database of posed images or 3D models. This is typically done by estimating 6-degree-of-freedom (6-DoF) camera poses within a world coordinate system defined by the database.

Traditional visual localization approaches rely on structure-from-motion (SfM) techniques~\cite{schonberger2016structure,detone2018superpoint,sarlin2020superglue} to reconstruct 3D models. These methods~\cite{sattler2016efficient,panek2022meshloc,dusmanu2019d2,revaud2019r2d2,sarlin2020superglue,sun2021loftr,lindenberger2023lightglue,wang2024efficient,giang2024learning,wang2024dgc} match pixels from query images to 3D scene points, then use geometric optimization~\cite{fischler1981random,hartley2003multiple,kneip2011novel,sarlin2019coarse,sarlin2021back} to solve camera poses. While renowned for their high localization accuracy, these approaches often struggle with inefficiencies at test time, which limits their scalability for real-time applications.
Scene coordinate regression approaches~\cite{brachmann2017dsac,brachmann2018learning,li2018full,li2020hierarchical,brachmann2021visual,dong2022visual,brachmann2023accelerated} provide an alternative perspective on pixel-to-point correspondences. These methods use neural networks to learn implicit scene representations, which are then used to infer dense correspondences. However, most of them face limitations in generalization. Moreover, these methods often require intensive supervision, such as ground-truth keypoint matches or 3D point maps, making it challenging to scale up the training data.

Absolute pose regression (APR) approaches~\cite{kendall2016modelling,walch2017image,kendall2017geometric,naseer2017deep,wu2017delving,melekhov2017image,brahmbhatt2018geometry,wang2020atloc,shavit2021learning,yin2022fishermatch,shavit2023coarse} directly regress camera poses from images, offering much faster inference times and high accuracy. However, most of these methods are inherently scene-specific and typically require dense viewpoint coverage during training, limiting their real-world applicability. Recent attempts~\cite{moreau2022lens,chen2021direct,chen2022dfnet,lin2024learning,chen2024neural} to enhance accuracy through synthetic data generation introduce significant computational overhead, hampering widespread deployment. 
In contrast, relative pose regression (RPR) approaches~\cite{laskar2017camera, balntas2018relocnet,en2018rpnet,abouelnaga2021distillpose,arnold2022map,laskar2017camera,zhou2020learn,winkelbauer2021learning} estimate relative poses between database-query image pairs. These methods mitigate the need for per-scene training while maintaining the test-time efficiency of APR models.
However, even the most advanced RPR methods~\cite{ding2019camnet,turkoglu2021visual,winkelbauer2021learning,arnold2022map} have not yet matched APR approaches in localization accuracy. While several RPR methods~\cite{laskar2017camera,turkoglu2021visual,winkelbauer2021learning,arnold2022map} have shown the ability to generalize across different datasets, this often results in further reduced camera pose accuracy. 
Therefore, most of the aforementioned approaches struggle with one of the following three criteria: novel-scene generalization, test-time efficiency, and camera pose accuracy.

To address these challenges, we present \textbf{Reloc3r} (pronounced ``\textipa{rI"loUk@r}''), a simple yet surprisingly effective visual localization framework. It draws inspiration from recent foundation models~\cite{kirillov2023segment,peebles2023scalable,liu2024visual,bai2024sequential,wang2024dust3r,leroy2024grounding,tong2024cambrian,wang2024moge}. These models, utilizing scalable network architectures (such as Transformers~\cite{dosovitskiy2020image,he2022masked}) and large-scale training, demonstrate strong performance and exceptional generalization across various tasks. This success motivates our exploration of a similar methodology for pose estimation. We adopt architecture from DUSt3R~\cite{wang2024dust3r} as the backbone, applying parsimonious and elegant modifications to build a relative pose regression network. The network is designed to be fully symmetric and disregards the metric scale of relative poses during training. We then integrate this with a minimalist motion averaging module to estimate absolute poses, resulting in the Reloc3r framework. To unleash the power of large-scale training, we processed around eight million image pairs from diverse public sources, spanning object-centric, indoor, and outdoor scenes. Experimentally, Reloc3r demonstrates superior performance across six well-known pose estimation datasets, benefiting from its simple architecture and large-scale training. Our key contributions can be summarized as follows:
\begin{itemize}

\item 
We introduce Reloc3r, a simple yet surprisingly effective visual localization framework. It enables excellent generalization to novel scenes, fast test-time efficiency, and high camera pose accuracy. 

\item 
Both the proposed fully symmetric relative pose regression network and the motion averaging module follow the principle of parsimony. This streamlined approach enables efficient large-scale training. 

\item 
Comprehensive experiments across six popular evaluation datasets consistently demonstrate the effectiveness of our proposed approach.

\end{itemize}

\section{Related Work}
\label{sec:related}

\myParagraph{Structure-based visual localization.} 
The structure-based localization pipeline is a well-established approach to solving camera pose via multi-view geometry~\cite{hartley2003multiple}. Modern methods~\cite{li2010location,sattler2011fast,sattler2012improving,zeisl2015camera,sattler2016efficient,sattler2017large,taira2018inloc,sattler2018benchmarking,humenberger2020robust,zhou2020learn,sarlin2021back,panek2022meshloc,yang2022scenesqueezer,giang2024learning,wang2024dgc,liu2025robust} typically consist of two main steps: 1) establishing correspondences either between images or between pixels and a pre-built 3D model, and 2) robustly solving the camera pose from noisy correspondences.
These correspondences are obtained through keypoint matching~\cite{lowe1999object,arandjelovic2012three,detone2018superpoint,dusmanu2019d2,revaud2019r2d2,sarlin2020superglue,sun2021loftr,wang2024efficient,lindenberger2023lightglue,edstedt2024roma,leroy2024grounding} or scene coordinate regression~\cite{brachmann2017dsac,brachmann2018learning,li2018full,yang2019sanet,li2020hierarchical,brachmann2021visual,tang2021learning,dong2022visual,brachmann2023accelerated, wang2024hscnet++}. Robust estimators~\cite{fischler1981random,chum2003locally,lebeda2012fixing,barath2018graph,barath2019magsac,barath2020magsac++,barath2021graph,PoseLib} are then applied to estimate the final camera pose. Though effective, these methods often have slow inference times, which limits their use in real-time applications.
Recent efficiency-oriented variants~\cite{lindenberger2023lightglue, wang2024efficient} have been proposed to speed up the matching process. However, the complex system design and high computational cost of robust estimation remain bottlenecks. These methods typically require ground-truth correspondences or 3D point maps for supervision, limiting their scalability for large-scale training. To address these challenges, we choose a straightforward camera pose regression approach. This method allows for efficient large-scale training while simplifying the system, resulting in a faster and more scalable visual localization solution. 

\myParagraph{Camera pose regression.} 
End-to-end pose regression~\cite{kendall2015posenet,kendall2016modelling,walch2017image,kendall2017geometric,naseer2017deep,wu2017delving,melekhov2017image,brahmbhatt2018geometry,wang2020atloc,shavit2021learning,yin2022fishermatch,shavit2023coarse,sattler2019understanding,moreau2022lens,chen2021direct,chen2022dfnet,lin2024learning,chen2024neural,chen2024map,laskar2017camera,balntas2018relocnet,en2018rpnet,abouelnaga2021distillpose,arnold2022map,zhou2020learn,winkelbauer2021learning,saha2018improved,ding2019camnet,turkoglu2021visual} has gained popularity due to its real-time inference capabilities. These methods fall into two broad categories: absolute pose regression (APR) and relative pose regression (RPR).

APR approaches directly regress camera positions and orientations in the world coordinate system from images within milliseconds. Despite their simplicity, these methods fall short of the localization accuracy achieved by structure-based approaches and often resemble pose approximation through image retrieval~\cite{sattler2019understanding}. To improve the accuracy of APR, some methods~\cite{moreau2022lens,chen2022dfnet,chen2024neural,lin2024learning} resort to novel view synthesis~\cite{mildenhall2021nerf} to create dense viewpoints for training. While being effective, this strategy introduces significant computational costs, with training taking hours or days for each specific scene.
More recent approach~\cite{chen2024map} attempts to reduce training time to minutes by building the connection between pose regressor and scene-specific maps. 
However, it is still limited to per-scene training and evaluation.

RPR approaches aim to generalize across different scenes by learning the relative pose between image pairs. Localization is achieved by regressing the relative pose between the query image and the most similar (or top-$K$) database images. The metric scale of the relative translation can be estimated approximately from a single database-query pair~\cite{balntas2018relocnet,en2018rpnet,abouelnaga2021distillpose,arnold2022map}, while more precise absolute positioning is possible through multi-view triangulation~\cite{laskar2017camera,zhou2020learn,winkelbauer2021learning,dong2023lazy}. Despite this capability, the best RPR methods~\cite{ding2019camnet,zhou2020learn,winkelbauer2021learning,arnold2022map,khatib2022leveraging} lag significantly behind APR methods. Furthermore, the generalization ability of existing RPR models remains limited. For instance, methods like Relative PN~\cite{laskar2017camera} and Relpose-GNN~\cite{turkoglu2021visual} can adapt to new datasets and scenes, but their localization errors nearly double in these scenarios.
Map-free~\cite{arnold2022map}, despite being trained on large datasets (about 523K samples), still relies on separate models for indoor and outdoor settings. 
Other approaches explore multi-view pose estimation~\cite{xue2020learning,zhang2022relpose,wang2023posediffusion,lin2024relpose++,zhang2024cameras}, as well as wide-baseline~\cite{rockwell2024far} and panoramic methods~\cite{tu2024panopose}. 
However, they remain dataset-specific training, which limits their scalability. 

Previous methods have largely focused on technical design rather than scaling training on larger and more diverse datasets. In this paper, we present the first pose regression approach trained on a diverse mixture of object-centric, indoor, and outdoor datasets. 
Contrary to prior work~\cite{zhou2020learn} suggesting that pose regression inaccuracies stem from coarse feature localization, we find that a well-trained patch-level regression network can achieve, and sometimes even surpass the performance of pixel-level feature matching.

\myParagraph{Foundation Models.} 
Large neural network models have advanced significantly due to the scalability of the Transformer architecture~\cite{vaswani2017attention}. This architecture underpins both large language models~\cite{devlin2018bert,ouyang2022training,touvron2023llama,chowdhery2023palm,bai2023qwen,minaee2024large} and large vision models~\cite{dosovitskiy2020image,kirillov2023segment,peebles2023scalable,bai2024sequential,ravi2024sam,liu2024visual,tong2024cambrian}. These foundation models, trained on large-scale datasets, have proven effective and demonstrate strong generalization capabilities across diverse tasks. Recently, DUSt3R~\cite{wang2024dust3r} introduces the first 3D foundation model capable of addressing nearly all tasks related to two-view geometry. Its architecture has been adapted and fine-tuned in numerous follow-up works~\cite{leroy2024grounding,smart2024splatt3r,wang20243d,zhang2024monst3r,ye2024no} to enhance performance in various downstream tasks. In our work, we adopt DUSt3R's Transformer backbone to develop Reloc3r, but featuring a fully symmetric design. We will discuss the benefits of this symmetric architecture in the following section.

\section{Method}
\label{sec:method}

\begin{figure*}[!t]
\centering
\includegraphics[width=0.99\linewidth]{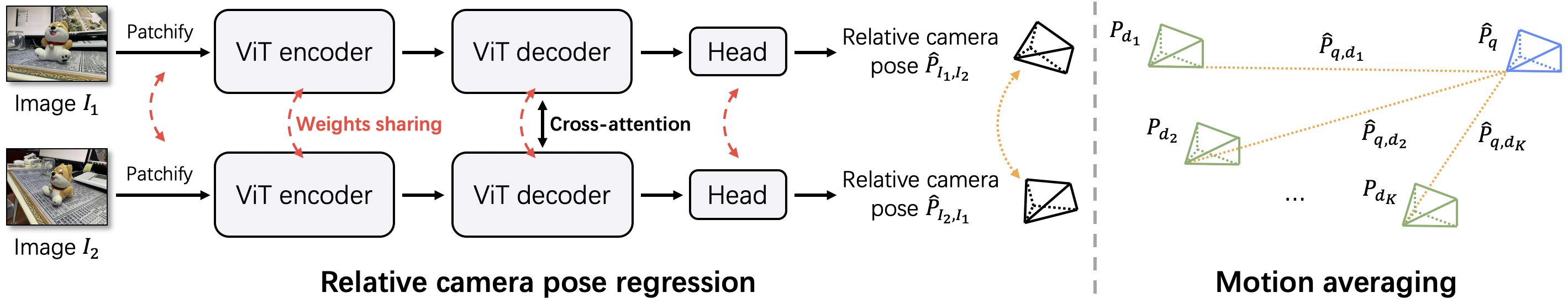}
\caption{
Reloc3r consists of two modules: a relative camera pose regression network (Sec.~\ref{sec:rpr}) and a motion averaging module (Sec.~\ref{sec:ma}).
Given a pair of input images, the network module infers the relative camera pose (at an unknown scale) between them. This module consists of a two-branch Vision Transformer (ViT) with shared weights. The images are divided into patches, converted to tokens, and embedded as latent features through separate encoders. Decoders then exchange information between the two sets of latent features. Each head aggregates its latent features to estimate a relative camera pose. To determine the absolute camera pose of a query image relative to a database, we retrieve at least two database-query pairs. These pairs are first processed by the network for relative pose estimation. Subsequently, the motion averaging module computes the absolute metric pose by aggregating the relative estimates.
}
\label{fig:overview}
\end{figure*}

\myParagraph{Problem statement.} 
Given a database $\textbf{D} = \{I_{d_n} \in \mathbb{R}^{H\times W \times 3} \ | \ n = 1, ..., N \}$ of posed images from a scene, and a query image $I_q$ from the same scene, the task of visual localization is to estimate the 6-DoF camera pose $P \in \mathbb{R}^{3 \times 4}$ that can register $I_q$ to the world coordinate system defined by the database images. $P$ is represented by camera rotation $R \in \mathbb{R}^{3 \times 3}$ and translation $t \in \mathbb{R}^3$.

\myParagraph{Method overview.}
Figure~\ref{fig:overview} illustrates an overview of the proposed visual localization method. It comprises two main components: a \textit{relative pose regression network}, and a \textit{motion averaging module}. The query image $I_q$ is paired with the top-$K$ database images using an off-the-shelf image retrieval approach~\cite{arandjelovic2016netvlad}, creating a set of image pairs $\textbf{Q} = \{(I_{d_k}, I_q) \ | \ k = 1,...,K \}$. The relative pose regression network processes each image pair from $\textbf{Q}$ independently to determine their relative poses $P_{d,q}$ and $P_{q,d}$. 
With known poses of database images $\{\hat{P}{d_1}, ..., \hat{P}{d_K}\}$, it is possible to infer the query image's absolute rotation from a single pair. However, these estimates can be noisy, and the translation vector's metric scale remains uncertain. To tackle these issues, a motion averaging module performs rotation averaging and camera center triangulation, eventually producing the absolute metric pose of the query image. 

In Sec.~\ref{sec:rpr}, we provide a detailed description of the pose regression model. This model leverages a unified Vision Transformer (ViT)~\cite{dosovitskiy2020image} architecture with minimal modifications. The motion averaging module discussed in Sec.~\ref{sec:ma} also takes a minimalist approach. It performs simple rotation averaging and camera center triangulation, without any trainable parameters involved.

\subsection{Relative Camera Pose Regression}
\label{sec:rpr}
The proposed relative camera pose regression network, inspired by DUSt3R~\cite{wang2024dust3r}, takes an image pair $I_1, I_2$ as input. While we assume the images have the same resolution (consistent with DUSt3R), their resolutions may differ in practice. The network divides the images into patches and processes them as tokens through a Vision Transformer (ViT) encoder. A ViT decoder~\cite{weinzaepfel2022croco,weinzaepfel2023croco} then uses a cross-attention mechanism to exchange information between tokens from both branches. This is followed by regression heads that predict the relative poses $\hat{P}_{I_1, I_2}$ and $\hat{P}_{I_2, I_1}$. The two branches are fully symmetrical with shared weights.

\myParagraph{ViT encoder-decoder architecture.}
The two modules resemble those of DUSt3R and gain substantial benefits from pre-training. We begin by dividing each input image $I_i$ into a sequence of $T$ tokens, each with dimension $d$. Next, we compute RoPE positional embeddings~\cite{su2024roformer} for each token to encode their relative spatial positions within the image. Then, we process the tokens through $m$ ViT encoder blocks, each comprising self-attention and feed-forward layers, to produce encoded feature tokens $F_1$ and $F_2$:
\begin{equation*}
F_{i}^{(T \times d)} = \text{Encoder}(\text{Patchify}(I_i^{(H \times W \times 3)})), \ \text{where } i=1, 2.
\end{equation*}

The decoder comprises $n$ ViT decoder blocks, each using the same RoPE position embedding. Unlike the encoder block, each decoder block incorporates an additional cross-attention layer between its self-attention and feed-forward layers. This structure enables the model to reason the spatial relationship between two sets of feature tokens.
We obtain decoded tokens as:
\begin{align*} 
G_{1}^{(T \times d)} &= \text{Decoder}(F_{1}^{(T \times d)}, F_{2}^{(T \times d)}) , \\
G_{2}^{(T \times d)} &= \text{Decoder}(F_{2}^{(T \times d)}, F_{1}^{(T \times d)}) .
\end{align*}

\myParagraph{Pose regression head.} 
Following recent research~\cite{chen2024map}, our pose regression head comprises $h$ feed-forward layers followed by average pooling, with additional layers to regress relative rotation and translation. The rotation is initially represented using a 9D representation~\cite{levinson2020analysis}. It is then converted into a 3$\times$3 rotation matrix using SVD orthogonalization. This matrix is concatenated with the 3D translation vector to form the final transformation matrix.
The final output for the relative pose is: 
\begin{equation*}
\hat{P}_{I_1, I_2}^{(3 \times 4)} = \text{Head}(G_{1}^{(T \times d)}), \ \ \ 
\hat{P}_{I_2, I_1}^{(3 \times 4)} = \text{Head}(G_{2}^{(T \times d)}) .
\end{equation*}

\myParagraph{Supervision signal.}
The predicted pose by our network conveys two pieces of information: 1) rotation, which measures the relative change in orientation, and 2) translation, which indicates the relative direction of camera center movement. Both of these quantities can be expressed as relative angles. Consequently, we train the network by minimizing the difference between these predicted relative angles and their ground-truth: 
$\mathcal{L} = \ell_R + \ell_t$, where 
\begin{equation*}
\ell_R = \text{arccos} ( \frac{\text{tr}(\hat{R}^{-1} R) - 1}{2} ), 
\ell_t = \text{arccos} ( \frac{\hat{t} \cdot t}{\lVert \hat{t} \rVert \lVert t \rVert} ).
\end{equation*}
Here, tr($\cdot$) denotes the trace of a matrix. $\hat{R}$ and $\hat{t}$ represent the predicted rotation and translation, while $R$ and $t$ represent their respective ground-truth values.

\myParagraph{Discussion.} 
Unlike DUSt3R's asymmetric branches for coordinate alignment, we use a fully symmetric architecture, which is inherently more suitable for relative pose estimation. This design eliminates biases from image ordering, thereby simplifying training. It also allows weight sharing across branches, reducing computational complexity and storage requirements.

Recent works~\cite{winkelbauer2021learning,arnold2022map} on relative pose regression prefer learning a metric pose. In contrast, we choose to learn only the direction of translation, as motion averaging can effectively solve the metric scale (discussed in the next section). This approach represents translations as angles in the same dimension as rotations, eliminating the need to weight rotation and translation values during training. It also avoids the challenge of balancing different dataset metric scales.

\subsection{Motion Averaging}
\label{sec:ma}

To maintain efficiency and simplicity, we integrate our pose regression network with a minimalist motion averaging module. Given the high accuracy of the network's predictions, we don't apply any robust estimation~\cite{fischler1981random,chatterjee2013efficient,wilson2014robust,zhuang2018baseline,dong2023lazy}. For each image pair, the regression network produces two relative poses: $\hat{P}_{I_1, I_2}$ and $\hat{P}_{I_2, I_1}$, which ideally should be inverses of each other. Empirically, we observe similar accuracy for both poses. By default, we use the transformation that maps the query to the database as input for motion averaging. The module processes rotation and translation separately, as detailed below.

\myParagraph{Rotation avgeraging.}
Given a relative rotation estimation $\hat{R}_{q, d_i}$ from a database-query pair, the absolute rotation is computed as $\hat{R}_q = R_{d_i} \hat{R}_{q, {d_i}}$. The motion averaging module reduces prediction noise by aggregating absolute rotation estimates from all available pairs. This aggregation is efficiently performed by calculating the mean rotation using quaternion representations~\cite{zhou2020learn}. We have observed that calculating the median rotation can further enhance robustness against noise, with minimal additional computational cost. Consequently, we use the median rotation as our final rotation estimation.

\myParagraph{Camera center triangulation.}
The absolute camera center can be triangulated~\cite{hartley1997triangulation} from two database-query pairs. Similar to rotation averaging, we calculate an average intersection using all valid pairs. While the geometric median of the intersection is not analytically solvable and typically requires iterative optimization, we opt for a more efficient approach. 
We use a simple least-squares method that minimizes the sum of squared distances from the camera center to each translation direction derived from relative pose estimates. The solution is obtained through Singular Value Decomposition (SVD).

\section{Experiments}
\label{sec:exp}

In this section, we first introduce our training data and implementation details, then evaluate our method across multiple datasets. In Sec.~\ref{sec:relpose}, we verify our network's performance on the relative pose estimation task. In Sec.~\ref{sec:visloc}, we evaluate our full method on the visual localization task. Finally, we conduct ablation studies and analyses in Sec.~\ref{sec:ablation}. 
It's worth noting that our method is trained once for all experiments, meaning all evaluations use the same model, except for the ablation studies. 

\myParagraph{Training data.} 
To build up-to-scale training pairs with ground-truth relative poses, similar to DUSt3R~\cite{wang2024dust3r}, we process around eight million image pairs from seven public datasets. 
The statistics are reported in Table~\ref{tab:train}.
We convert the ground-truth relative poses to a unified format: relative coordinate transformation aligned with the OpenCV~\cite{bradski2000opencv} definition. Each image is center-cropped based on its principal points and resized to a width of 512 pixels.

\begin{table}[th!]
\centering
\resizebox{0.48\textwidth}{!}{

\small

\begin{tabular}{l|cc}
\toprule
Datasets & Scene type & Number of image pairs \\

\toprule

CO3Dv2~\cite{reizenstein2021common} & Object-centric & $\sim$1M \\
ScanNet++~\cite{yeshwanth2023scannet++} & Indoor & $\sim$850K \\ 
ARKitScenes~\cite{baruch2021arkitscenes} & Indoor & $\sim$2.1M \\
BlendedMVS~\cite{yao2020blendedmvs} & Outdoor & $\sim$1M \\ 
MegaDepth~\cite{li2018megadepth} & Outdoor & $\sim$1.8M \\ 
DL3DV~\cite{ling2024dl3dv} & Indoor \& outdoor & $\sim$1.1M \\ 
RealEstate10K~\cite{zhou2018stereo} & Indoor \& outdoor & $\sim$100K \\ 

\bottomrule

\end{tabular}
}
\caption{
Training data of Reloc3r. It comprises about eight million image pairs from seven public sources, covering a wide range of scenes from object-centric to indoor and outdoor environments.
}
\label{tab:train}
\end{table}

\myParagraph{Implementation details.} 
By default, the proposed Reloc3r uses $m=24$ encoder blocks, $n=12$ decoder blocks, followed by the pose regression head with $h=2$ convolutional layers. To improve memory usage and speed in self and cross-attention, we employ memory-efficient attention. This design achieves approximately a 14\% speed increase while saving 25\% of GPU memory during training. We initialize Reloc3r with DUSt3R's pre-trained 512-DPT weights. For decoder initialization, we use the weights from DUSt3R decoder2, as it's pre-trained to perform coordinate transformation. The full model is trained on 8 AMD MI250x-40G GPUs with a batch size of 8 and a learning rate starting at 1e-5, decaying to 1e-7.

For the visual localization task, following the literature~\cite{sarlin2019coarse, turkoglu2021visual}, we apply NetVLAD~\cite{arandjelovic2016netvlad} for image retrieval and use the top 10 similar image pairs. We directly use these retrieved image pairs without distance-based clustering, filtering, or other heuristics to enhance their spatial distribution. All evaluations are conducted on a 24GB NVIDIA GeForce RTX 4090 GPU and we use mixed precision fp16 / fp32 to improve speed and memory efficiency without sacrificing accuracy.

\subsection{Relative Camera Pose Estimation}
\label{sec:relpose}

\begin{table}[t!]
\centering
\resizebox{0.49\textwidth}{!}{

\small

\begin{tabular}{c|l|ccc}
\toprule
& Methods & RRA@15 & RTA@15 & mAA@30 \\

\toprule

\multirow{8}{*}{\rotatebox{90}{Non-PR}} 

& PixSfM~\cite{lindenberger2021pixel} & 33.7 & 32.9 & 30.1  \\

& RelPose~\cite{zhang2022relpose} & 57.1 & - & -  \\

& PoseDiffusion~\cite{wang2023posediffusion} & 80.5 & 79.8 & 66.5 \\

& RelPose++~\cite{lin2024relpose++} & 82.3 & 77.2 & 65.1 \\

& RayDiffusion*~\cite{zhang2024cameras} & 93.3 & - & - \\

& VGGSfM~\cite{wang2024vggsfm} & 92.1 & 88.3 & 74.0 \\

& 
DUSt3R (w/ PnP)~\cite{wang2024dust3r} & 94.3 & 88.4 & 77.2 \\

&
MASt3R~\cite{leroy2024grounding} 
& \textbf{94.6} 
& \textbf{91.9} 
& \textbf{81.1} \\

\midrule

\multirow{4}{*}{\rotatebox{90}{PR}} 

& PoseReg~\cite{wang2023posediffusion} & 53.2 & 49.1 & 45.0 \\ 

& RayReg*~\cite{zhang2024cameras} & 89.2 & - & - \\

& \textbf{Reloc3r-224 (Ours)} 
& 93.6
& 91.9
& 79.1\\

& \textbf{Reloc3r-512 (Ours)} 
& \underline{\textbf{95.8}}
& \underline{\textbf{93.7}} 
& \underline{\textbf{82.9}} \\

\bottomrule
\end{tabular}
}
\caption{
Relative pose evaluation (multi-view) on the CO3Dv2 dataset~\cite{reizenstein2021common}. 
The best results for each method category are highlighted in bold. 
The underlined numbers indicate that our method achieves the best performance among all competitors. The methods marked with * donate evaluation on 8 frames. 
}
\label{tab:co3d}
\end{table}

\begin{table*}[t!]
\centering

\resizebox{1.0\textwidth}{!}{

\small

\begin{tabular}{c|l|ccc|ccc|ccc|c}
\toprule
& \multirow{2}{*}{Methods} & \multicolumn{3}{|c|}{ScanNet1500} & \multicolumn{3}{|c|}{RealEstate10K} & \multicolumn{3}{|c|}{ACID} & \multirow{2}{*}{Inference time} \\
& & AUC@5 & AUC@10 & AUC@20 & AUC@5 & AUC@10 & AUC@20 & AUC@5 & AUC@10 & AUC@20 \\
\toprule

\multirow{5}{*}{\rotatebox{90}{Non-PR}} 

& Efficient LoFTR~\cite{wang2024efficient}
& 19.20 & 37.00 & 53.60 & - & - & - & - & - & - & {40 ms} \\

& ROMA~\cite{edstedt2024roma} & 28.90 & 50.40 & 68.30 & 54.60 & 69.80 & 79.70 & 46.30 & 58.80 & 68.90 & 300 ms \\ 

& DUSt3R~\cite{wang2024dust3r} & 23.81 & 45.91 & 65.57 & 39.70 & 56.88 & 70.43 & 21.50 & 35.95 & 49.70 & 441 ms \\

& MASt3R~\cite{leroy2024grounding} & 28.01 & 50.24 & 68.83 & 63.54 & 76.39 & 84.50 & \textbf{52.12} & \textbf{64.54} & \textbf{73.61} & 294 ms \\

& NoPoSplat~\cite{ye2024no} & \textbf{31.80} & \textbf{53.80} & \textbf{71.70} & \textbf{69.10} & \textbf{80.60} & \textbf{87.70} & {48.60} & {61.70} & {72.80} & $>$2000 ms \\ 

\midrule

\multirow{6}{*}{\rotatebox{90}{PR}} 

& {Map-free (Regress-SN)}~\cite{arnold2022map} 
& 1.84 & 8.75 & 25.33 
& 0.83 & 4.06 & 13.97
& 1.32 & 5.82 & 16.28
& {10 ms} \\

& {Map-free (Regress-MF)}~\cite{arnold2022map}
& 0.50 & 3.48 & 13.15 
& 1.61 & 6.74 & 18.38
& 2.57 & 9.96 & 24.50
& {10 ms} \\

& {ExReNet (SN)}~\cite{winkelbauer2021learning}
& 2.30 & 10.71 & 26.13 
& 2.17 & 7.94 & 20.43 
& 1.90 & 7.53 & 18.69
& {17 ms} \\

& {ExReNet (SUNCG)}~\cite{winkelbauer2021learning}
& 1.61 & 7.00 & 18.03 
& 3.27 & 12.06 & 27.85
& 4.14 & 13.43 & 27.70
& {17 ms} \\ 

& \textbf{Reloc3r-224 (Ours)} 
& 28.34
& 52.60
& 71.56
& 59.70
& 75.05
& 84.71
& 28.25
& 47.34
& 62.54
& 15 ms \\

& \textbf{Reloc3r-512 (Ours)} 
& \underline{\textbf{34.79}} 
& \underline{\textbf{58.37}} 
& \underline{\textbf{75.56}} 
& \textbf{66.70} 
& \textbf{80.20} 
& \underline{\textbf{88.39}}
& \textbf{38.18} 
& \textbf{56.39} 
& \textbf{70.34} 
& 25 ms \\

\bottomrule
\end{tabular}

}
\caption{
Relative camera pose evaluation on the ScanNet1500~\cite{dai2017scannet,sarlin2020superglue}, RealEstate10K~\cite{zhou2018stereo} and ACID~\cite{liu2021infinite} datasets. 
The best results for each method category are highlighted in bold. 
Our method outperforms all the pose regression competitors. Moreover, it achieves the best results (underlined) among all the competitors on several datasets and metrics. Notably, our method runs in real-time, which is over $50\times$ faster than the state-of-the-art Non-PR method. 
{Please refer to Figure~\ref{fig:teaser} for an intuitive comparison with representative methods. }
}
\label{tab:scannet}
\end{table*}

\begin{table*}[t!]
\centering
\resizebox{1.0\textwidth}{!}{

\small

\begin{tabular}{c|l|ccccccc|c|c}
\toprule
& \multirow{2}{*}{Methods} & \multirow{2}{*}{Chess} & \multirow{2}{*}{Fire} & \multirow{2}{*}{Heads} & \multirow{2}{*}{Office} & \multirow{2}{*}{Pumpkin} & \multirow{2}{*}{RedKitchen} & \multirow{2}{*}{Stairs} & \multirow{2}{*}{Average} & Dataset-specific \\
&  &  &  &  &  &  &  &  &  & training time \\

\toprule

\multirow{4}{*}{\rotatebox{90}{APR}} 

& LENS~\cite{moreau2022lens} & 0.03 / 1.30 & 0.10 / 3.70 & 0.07 / 5.80 & 0.07 / 1.90 & 0.08 / 2.20 & 0.09 / 2.20 & 0.14 / 3.60 & 0.08 / 3.00 & Days / scene \\

& PMNet~\cite{lin2024learning} & 0.03 / 1.26 & 0.04 / 1.76 & \textbf{0.02} / 1.68 & 0.06 / 1.69 & 0.07 / 1.96 & 0.08 / 2.23 & 0.11 / 2.97 & 0.06 / 1.93 & Days / scene \\

& DFNet~\cite{chen2022dfnet}+NeFeS~\cite{chen2024neural} & \textbf{0.02 / 0.57} & \textbf{0.02 / 0.74} & \textbf{0.02 / 1.28} & \textbf{0.02 / 0.56} & \textbf{0.02 / 0.55} & \textbf{0.02 / 0.57} & \textbf{0.05 / 1.28} & \textbf{0.02 / 0.79} & Days / scene \\

& Marepo~\cite{chen2024map} & \textbf{0.02} / 1.24 & \textbf{0.02} / 1.39 & \textbf{0.02} / 2.03 & 0.03 / 1.26 & 0.04 / 1.48 & 0.04 / 1.71 & 0.06 / 1.67 & 0.03 / 1.54 & 15min / scene \\

\midrule

\multirow{7}{*}{\rotatebox{90}{RPR (Seen)}} 

& {EssNet (7S)}~\cite{zhou2020learn} & - & - & - & - & - & - & - & 0.22 / 8.03 & Hours \\

& Relative PN (7S)~\cite{laskar2017camera} & 0.13 / 6.46 & 0.26 / 12.72 & 0.14 / 12.34 & 0.21 / 7.35 & 0.24 / 6.35 & 0.24 / 8.03 & 0.27 / 11.82 & 0.21 / 9.30 & Hours \\ 

& {NC-EssNet (7S)}~\cite{zhou2020learn} & - & - & - & - & - & - & - & 0.21 / 7.50 & Hours \\

& RelocNet (7S)~\cite{balntas2018relocnet} & 0.12 / 4.14 & 0.26 / 10.4 & 0.14 / 10.5 & 0.18 / 5.32 & 0.26 / 4.17 & 0.23 / 5.08 & 0.28 / 7.53 & 0.21 / 6.73 & Hours \\

& Relpose-GNN~\cite{turkoglu2021visual} & 0.08 / 2.70 & 0.21 / 7.50 & 0.13 / 8.70 & 0.15 / 4.10 & 0.15 / 3.50 & 0.19 / 3.70 & 0.22 / 6.50 & 0.16 / 5.20 & Hours \\

& AnchorNet~\cite{saha2018improved} & 0.06 / 3.89 & 0.15 / 10.3 & 0.08 / 10.9 & 0.09 / 5.15 & 0.10 / 2.97 & 0.08 / 4.68 & 0.10 / 9.26 & 0.09 / 6.74 & Hours \\

& CamNet~\cite{ding2019camnet} & \textbf{0.04 / 1.73} & \textbf{0.03 / 1.74} & \textbf{0.05 / 1.98} & \textbf{0.04 / 1.62} & \textbf{0.04 / 1.64} & \textbf{0.04 / 1.63} & \textbf{0.04 / 1.51} & \textbf{0.04 / 1.69} & Hours \\

\midrule

\multirow{11}{*}{\rotatebox{90}{RPR (Unseen)}} 

& {EssNet (CL)}~\cite{zhou2020learn} & - & - & - & - & - & - & - & 0.57 / 80.06 & None \\

& {NC-EssNet (CL)}~\cite{zhou2020learn} & - & - & - & - & - & - & - & 0.48 / 32.97 & None \\

& Relative PN (U)~\cite{laskar2017camera} & 0.31 / 15.05 & 0.40 / 19.00 & 0.24 / 22.15 & 0.38 / 14.14 & 0.44 / 18.24 & 0.41 / 16.51 & 0.35 / 23.55 & 0.36 / 18.38 & None \\ 

& RelocNet (SN)~\cite{balntas2018relocnet} & 0.21 / 10.9 & 0.32 / 11.8 & 0.15 / 13.4 & 0.31 / 10.3 & 0.40 / 10.9 & 0.33 / 10.3 & 0.33 / 11.4 & 0.29 / 11.3 & None \\

& ImageNet+NCM~\cite{zhou2020learn}$\dagger$ & - & - & - & - & - & - & - & 0.19 / 4.30 & None \\

& {Map-free (Match)}~\cite{arnold2022map}$\dagger$ & 
0.10 / 2.93 &
0.12 / 4.95 &
0.11 / 5.40 &
0.12 / 3.01 &
0.16 / 3.19 &
0.14 / 3.45 &
0.21 / 4.50 &
0.14 / 3.92 & None \\ 

& Map-free (Regress)~\cite{arnold2022map} & 0.09 / 2.66 &
0.13 / 4.54 &
0.11 / 4.81 &
0.11 / 2.77 &
0.16 / 3.11 &
0.14 / 3.48 &
0.18 / 4.70 &
0.13 / 3.72 & None \\ 

& {ExReNet (SN)}~\cite{winkelbauer2021learning} & 0.06 / 2.15 & 0.09 / 3.20 & 0.04 / 3.30 & 0.07 / 2.17 & 0.11 / 2.65 & 0.09 / 2.57 & 0.33 / 7.34 & 0.11 / 3.34 & None \\

& {ExReNet (SUNCG)}~\cite{winkelbauer2021learning} & 0.05 / 1.63 & 0.07 / 2.54 & 0.03 / 2.71 & 0.06 / 1.75 & 0.07 / 2.04 & 0.07 / 2.10 & 0.19 / 4.87 & 0.08 / 2.52 & None \\

& \textbf{Reloc3r-224 (Ours)} 
& \textbf{0.03} / 0.99
& 0.04 / 1.13
& 0.02 / 1.23
& 0.05 / \textbf{0.88}
& 0.07 / 1.14
& 0.05 / \textbf{1.23}
& 0.12 / 2.25
& 0.05 / 1.26
& None \\

& \textbf{Reloc3r-512 (Ours)} 
& \textbf{0.03 / 0.88} 
& \textbf{0.03 / 0.81} 
& \underline{\textbf{0.01 / 0.95}} 
& \textbf{0.04 / 0.88} 
& \textbf{0.06 / 1.10} 
& \textbf{0.04} / 1.26 
& \textbf{0.07 / \underline{1.26}} 
& \textbf{0.04 / 1.02} & None \\

\bottomrule
\end{tabular}

}
\caption{
Visual localization results on the 7 Scenes dataset~\cite{shotton2013scene}. 
We report median pose errors in meters and degrees. 
The best results for each method category are highlighted in bold. 
The underlined numbers indicate where Reloc3r outperforms all competitors. 
The methods marked with $\dagger$ indicate hybrid pose estimation, which combines additional geometric solvers with feature matching.
}
\label{tab:7s}
\end{table*}

\begin{table*}[t!]
\centering
\resizebox{1.0\textwidth}{!}{

\small

\begin{tabular}{c|l|ccccc|c|c|c}
\toprule
& \multirow{2}{*}{Methods} & \multirow{2}{*}{GreatCourt} & \multirow{2}{*}{KingsCollege} & \multirow{2}{*}{OldHospital} & \multirow{2}{*}{ShopFacade} & \multirow{2}{*}{StMarysChurch} & \multirow{2}{*}{Average (4)} & \multirow{2}{*}{Average} & Dataset-specific \\
&  &  &  &  &  &  &  &  & training time \\

\toprule

\multirow{3}{*}{\rotatebox{90}{APR}} 

& LENS~\cite{moreau2022lens} & - & 0.33 / \textbf{0.50} & \textbf{0.44} / 0.90 & 0.27 / 1.60 & 0.53 / 1.60 & 0.39 / 1.20 & - & Days / scene \\

& PMNet~\cite{lin2024learning} & - & \textbf{0.31} / 0.55 & \textbf{0.44 / 0.79} & 0.17 / 0.86 & \textbf{0.31 / 0.96} & \textbf{0.31} / 0.79 & - & Days / scene \\

& DFNet~\cite{chen2022dfnet}+NeFeS~\cite{chen2024neural} & - & 0.37 / 0.54 & 0.52 / 0.88 & \textbf{0.15 / 0.53} & 0.37 / 1.14 & 0.35 / \textbf{0.77} & - & Days / scene \\

\midrule

\multirow{4}{*}{\rotatebox{90}{RPR (Seen)}} 

& {EssNet (CL)}~\cite{zhou2020learn} & - & - & - & - & - & 1.08 / 3.41 & - & Hours \\

& Relpose-GNN~\cite{turkoglu2021visual} & 3.20 / 2.20 & \textbf{0.48} / 1.00 & \textbf{1.14 / 2.50} & \textbf{0.48} / 2.50 & 1.52 / 3.20 & 0.91 / 2.30 & 1.37 / 2.30 & Hours \\

& {NC-EssNet (CL)}~\cite{zhou2020learn} & - & - & - & - & - & 0.85 / 2.82 & - & Hours \\

& AnchorNet~\cite{saha2018improved} & - & 0.57 / \textbf{0.88} & {1.21 / 2.55} & 0.52 / \textbf{2.27} & \textbf{1.04 / 2.69} & \textbf{0.84 / 2.10} & - & Hours  \\

\midrule

\multirow{9}{*}{\rotatebox{90}{RPR (Unseen)}} 

& {EssNet (7S)}~\cite{zhou2020learn} & - & - & - & - & - & 10.36 / 85.75 & - & None \\ 

& {NC-EssNet (7S)}~\cite{zhou2020learn} & - & - & - & - & - & 7.98 / 24.35 & - & None \\ 

& {Map-free (Match)}~\cite{arnold2022map}$\dagger$ & 9.09 / 5.33 &
2.51 / 3.11 &
3.89 / 6.44 &
1.04 / 3.61 &
3.00 / 6.14 &
2.61 / 4.83 &
3.90 / 4.93 & None \\

& {Map-free (Regress)}~\cite{arnold2022map} & 8.40 / 4.56 &
2.44 / 2.54 &
3.73 / 5.23 &
0.97 / 3.17 &
2.91 / 5.10 &
2.51 / 4.01 & 
3.69 / 4.12 &
None \\

& {ExReNet (SN)}~\cite{winkelbauer2021learning} & 10.97 / 6.52 & 2.48 / 2.92 & 3.47 / 3.90 & 0.90 / 3.27 & 2.60 / 4.98 & 2.36 / 3.77 & 4.08 / 4.32 & None \\

& {ExReNet (SUNCG)}~\cite{winkelbauer2021learning} & 9.79 / 4.46 & 2.33 / 2.48 & 3.54 / 3.49 & 0.72 / 2.41 & 2.30 / 3.72 & 2.22 / 3.03 & 3.74 / 3.31 & None \\

& ImageNet+NCM~\cite{zhou2020learn}$\dagger$ & - & - & - & - & - & 0.83 / 1.36 & - & None \\

& \textbf{Reloc3r-224 (Ours)} 
& 1.71 / 0.94
& 0.47 / 0.41
& 0.87 / 0.66
& 0.18 / \underline{\textbf{0.53}}
& 0.41 / 0.73
& 0.48 / 0.58
& 0.73 / 0.65
& None \\

& \textbf{Reloc3r-512 (Ours)} 
& \underline{\textbf{1.22 / 0.73}}
& \textbf{0.42 / \underline{0.36}} 
& \textbf{0.62 / \underline{0.55}}  
& \textbf{\underline{0.13}} / 0.58 
& \textbf{0.34 / \underline{0.58}}  
& \textbf{0.38 / \underline{0.52}} & \underline{\textbf{0.55 / 0.56}} & None \\

\bottomrule
\end{tabular}

}
\caption{
Visual localization results on the Cambridge Landmarks~\cite{kendall2015posenet}. 
We report median pose errors in meters and degrees. 
The best results for each method category are highlighted in bold. 
The underlined numbers indicate where Reloc3r outperforms all competitors. 
The methods marked with $\dagger$ indicate hybrid pose estimation, which combines additional geometric solvers with feature matching. 
}
\vspace{-10pt}
\label{tab:cambridge}
\end{table*}

\begin{figure*}[!tbh]
\centering
\includegraphics[width=1\linewidth]{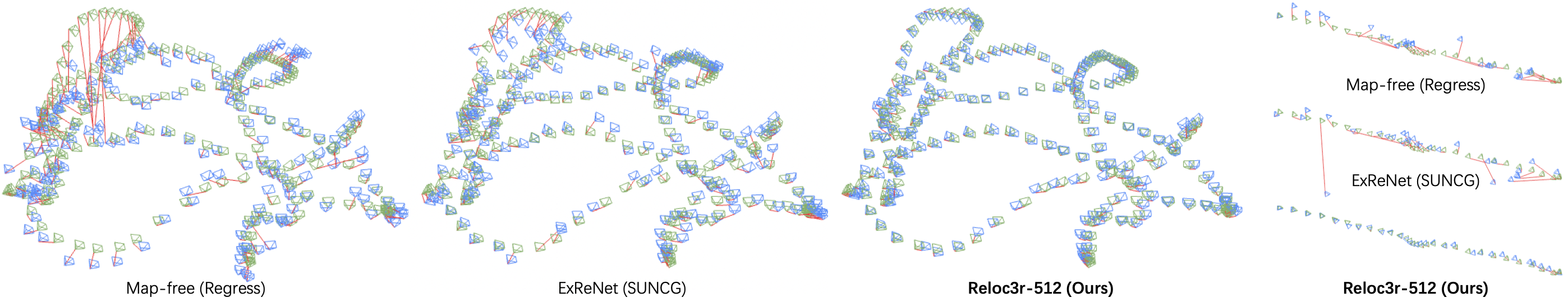}
\caption{
We visualize pose estimates for two scenes: Chess from the 7 Scenes dataset ~\cite{shotton2013scene} and KingsCollege from Cambridge Landmarks ~\cite{kendall2015posenet}. We compare Reloc3r's results with those of the most closely related RPR methods: ExReNet ~\cite{winkelbauer2021learning} and Map-free ~\cite{arnold2022map}. We can observe that Reloc3r's pose estimates align more closely with the ground-truth poses. 
}
\label{fig:visual}
\end{figure*}

In this section, we evaluate Reloc3r's relative pose regression module. Our evaluation covers two scenarios: pair-wise relative pose on ScanNet1500~\cite{dai2017scannet,sarlin2020superglue}, RealEstate10K~\cite{zhou2018stereo}, and ACID~\cite{liu2021infinite} datasets, as well as multi-view relative pose on Co3dv2~\cite{reizenstein2021common} dataset.

\myParagraph{Pair-wise relative pose.} 
Following recent research~\cite{ye2024no}, we evaluate pair-wise relative pose on three datasets: ScanNet1500 \cite{dai2017scannet,sarlin2020superglue}, RealEstate10K \cite{zhou2018stereo}, and ACID \cite{liu2021infinite}. These datasets showcase a variety of indoor and outdoor scenes captured with diverse camera trajectories. ScanNet1500~\cite{dai2017scannet,sarlin2020superglue} focuses on indoor scenes, RealEstate10K~\cite{zhou2018stereo} covers both indoor and outdoor environments, and ACID~\cite{liu2021infinite} is an aerial outdoor dataset. Importantly, our evaluation test sets do not overlap with scenes in our training data, and ACID remains entirely unseen by Reloc3r during training. Following previous research~\cite{sarlin2020superglue, sun2021loftr, wang2024efficient}, we employ three metrics: AUC@5/10/20. These metrics calculate the area under the curve of pose accuracy using thresholds of $\tau = 5/10/20$ degrees for the minimum of rotation and translation angular errors.

Since no existing pose regression (PR) methods have been specifically designed for these datasets, we primarily compare our approach with non-PR methods. Additionally, we evaluate the current state-of-the-art relative pose regression methods, ExReNet~\cite{winkelbauer2021learning,song2017semantic} and Map-free~\cite{arnold2022map}, on these datasets. 
The results are presented in Table~\ref{tab:scannet}. Reloc3r outperforms other PR methods by a significant margin across all three datasets. When compared to non-PR methods, 
Reloc3r also achieves SoTA performance. 
Specifically, on the ScanNet1500 dataset, Reloc3r delivers top performance with AUC at all thresholds, outperforming its baseline DUSt3R~\cite{wang2024dust3r} by around 13\% on AUC@20. 
On the RealEstate10K dataset, Reloc3r achieves performance comparable to NoPoSplat~\cite{ye2024no}, with AUC values of 66.70 / 80.20 / 88.39\% at thresholds of 5, 10, and 20, respectively. 
Similarly, on the ACID dataset, Reloc3r achieves high accuracy (AUC@20 = 70.34\%), demonstrating its generalization ability to unseen datasets. 
Furthermore, Reloc3r runs at an inference time of just 42 ms at an image resolution of 512 in width. This is significantly faster than many non-PR methods such as NoPoSplat~\cite{ye2024no} ($>$2000 ms) and ROMA~\cite{edstedt2024roma} (300 ms), 
and on par with PR methods. Such speed makes Reloc3r ideal for real-time applications.

\myParagraph{Multi-view relative pose.} 
We evaluate the multi-view relative pose on the Co3dv2~\cite{reizenstein2021common} dataset. This dataset consists of object-level scenes captured with inward-facing camera trajectories. The main challenges of this dataset include visual symmetries, textureless objects, and wide baselines between images. Following the evaluation protocol in~\cite{wang2023posediffusion, wang2024dust3r}, we evaluate Reloc3r on the test sets of 41 categories. For each sequence, we randomly sample 10 frames and formulate all 45 pairs for the evaluation. 
These relative poses are evaluated with three metrics: the relative rotation accuracy within 15 degrees (RRA@15), the relative translation accuracy within 15 degrees (RTA@15), and the mean average accuracy (mAA@30, also called AUC@30).

We compare Reloc3r with recent SfM-based approaches, PixSfM~\cite{lindenberger2021pixel} and VGGSfM~\cite{wang2024vggsfm}, as well as data-driven methods including RelPose~\cite{zhang2022relpose}, RelPose++~\cite{lin2024relpose++}, PoseDiffusion~\cite{wang2023posediffusion}, RayDiffusion~\cite{zhang2024cameras}, DUSt3R (with PnP)~\cite{wang2024dust3r}, and MASt3R~\cite{leroy2024grounding}. In addition, we also compare the regression-based methods PoseReg~\cite{wang2023posediffusion} and RayReg~\cite{zhang2024cameras}. 
Note that while we refer to our evaluation as multi-view, we actually only use pair-wise evaluation, similar to DUSt3R (w/ PnP) and MASt3R. In contrast, all other methods (excluding RayDiffusion and RayReg use 8 frames ) use all 10 frames simultaneously, thus having more contextual information for evaluation.

The quantitative results are reported in Table~\ref{tab:co3d}. Compared to existing approaches, the proposed Reloc3r achieves SoTA performance across multiple metrics, demonstrating significant improvements over competing methods. 
Specifically, Reloc3r achieves the best scores in terms of RRA@15 (95.8\%), RTA@15 (93.7\%), and mAA@30 (82.9\%), surpassing both Non-PR and PR methods.
This highlights Reloc3r's ability to robustly and accurately localize images in multi-view settings and wide baselines.

\subsection{Visual Localization}
\label{sec:visloc}

In this section, we evaluate Reloc3r’s absolute pose estimation for visual localization. We conduct experiments using two public datasets: the 7 Scenes dataset~\cite{shotton2013scene} and Cambridge Landmarks~\cite{kendall2015posenet}. The 7 Scenes dataset comprises seven indoor room scenes, each containing several video sequences captured from different moving trajectories. Cambridge Landmarks is a suburban-scale outdoor dataset featuring six scenes. Following prior approaches~\cite{brachmann2017dsac, li2020hierarchical,turkoglu2021visual}, we use five of these scenes for evaluation. For each scene, we report the median translation and rotation errors (in meters and degrees, respectively). It's worth noting that both datasets were entirely unseen by Reloc3r during training.

\myParagraph{Indoor visual localization.}
We compare Reloc3r with state-of-the-art absolute pose regression (APR) and relative pose regression (RPR) methods on the 7 Scenes~\cite{shotton2013scene} dataset. The results are shown in Table~\ref{tab:7s}. For RPR methods, we categorize them into two groups: \textit{seen}, where the model is trained and evaluated on the same dataset, and \textit{unseen}, where the dataset is entirely new to the model during evaluation. The table shows that methods like EssNet~\cite{zhou2020learn}, Relative PN~\cite{laskar2017camera}, and RelocNet~\cite{balntas2018relocnet} experience significant performance drops when evaluated on unseen datasets, revealing their limitations in scene generalizability. In contrast, our method consistently outperforms all RPR methods, even those trained on the same dataset, achieving an average median error of $0.04m/1.02^\circ$. Moreover, when compared with APR methods, our approach demonstrates comparable performance without requiring scene-specific training, further highlighting Reloc3r's robustness and adaptability across diverse scenes.

\myParagraph{Outdoor visual localization.} 
For the Cambridge~\cite{kendall2015posenet} dataset, a widely used benchmark for regression-based localization, all RPR methods (both \textit{seen} and \textit{unseen}) face challenges in this outdoor setting. Our proposed Reloc3r, as shown in Table~\ref{tab:cambridge}, surpasses all previous RPR methods without any retraining or fine-tuning on specific scenes, achieving consistent improvements across all evaluated scenes. Notably, in \textit{unseen} conditions, Reloc3r demonstrates an average halving of the pose error compared to the previous state-of-the-art RPR method, ImageNet+NCM~\cite{zhou2020learn}, with average errors of $0.38m/0.52^{\circ}$ across the last four scenes. Furthermore, our method shows a better average rotation error than all APR-based methods. This supports our motivation that leveraging a simple architectural design, coupled with scaled-up training, can effectively achieve good performance.
Visual comparisons are shown in Figure~\ref{fig:visual}.

\vspace{-1pt}

\subsection{Analyses}
\label{sec:ablation}

\vspace{-1pt}

\begin{table}[tb!]
\centering
\resizebox{0.43\textwidth}{!}{

\small

\begin{tabular}{l|ccc}
\toprule
\multirow{2}{*}{Methods} & \multicolumn{3}{|c}{ScanNet1500} \\
& AUC@5 & AUC@10 & AUC@20 \\
\toprule

Reloc3r-512 asymmetric & 32.71 & 56.84 & 74.63 \\ 

Reloc3r-512 metric pose & {25.70} & {50.20} & {70.07} \\

Reloc3r-512 (default) & \textbf{34.79} & \textbf{58.37} & \textbf{75.56} \\

\bottomrule
\end{tabular}
}
\caption{
Ablation studies examining asymmetric network architecture and pose prediction with metric scales. 
}

\vspace{-2pt} 

\label{tab:ablation}
\end{table}

\begin{figure}[!t]
\centering
\includegraphics[width=0.96\linewidth]{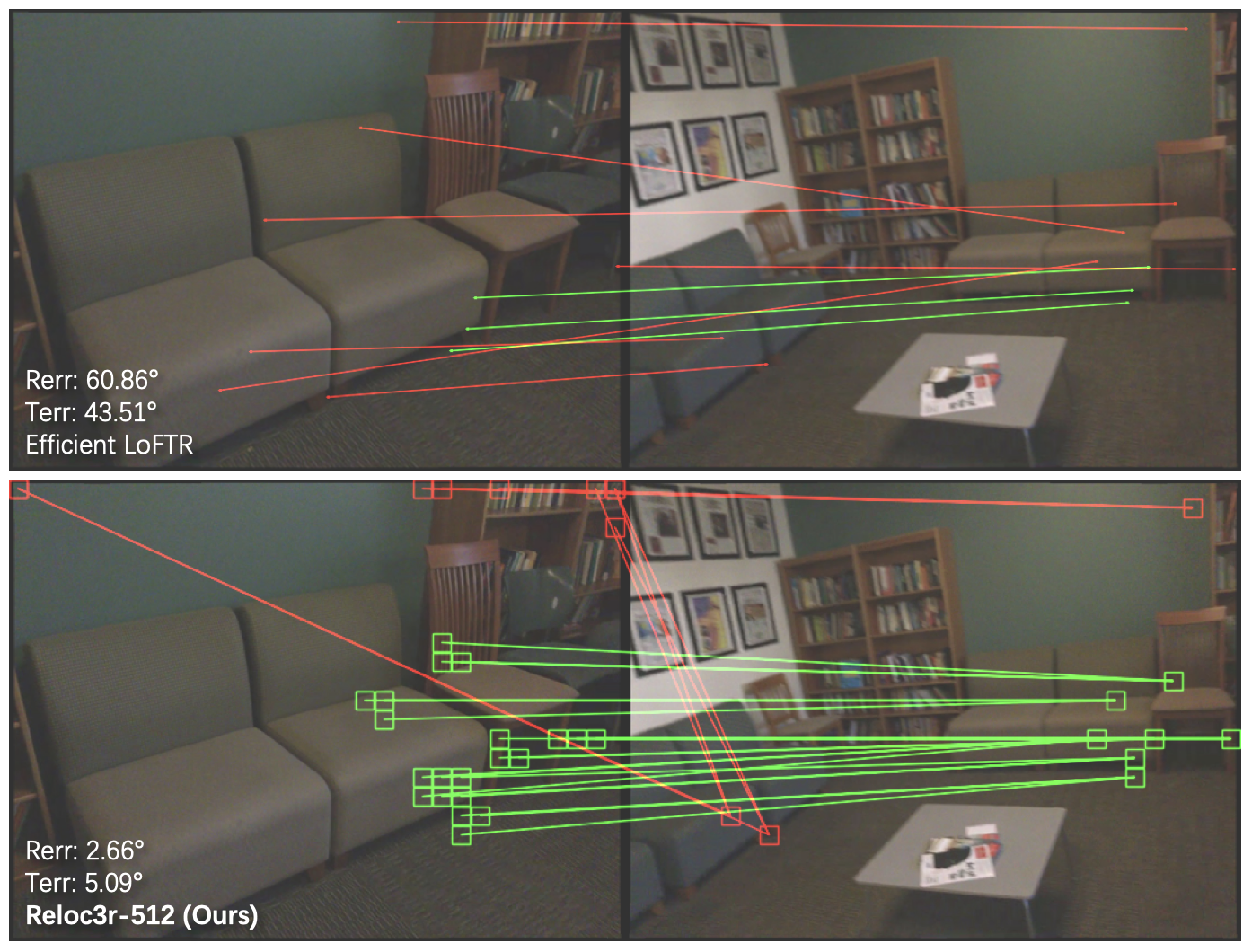}
\caption{
The top row showcases matches from Efficient LoFTR, while the bottom row displays the top-3 cross-attention responses from Reloc3r's decoder. We observe that the correlated regions in Reloc3r are superior to those of Efficient LoFTR, even though Reloc3r is trained solely with pose supervision.
}
\vspace{-2pt}
\label{fig:match}
\end{figure}

We summarize most insights here and refer the reader to the supplementary material for more details and discussions. 

\myParagraph{Image resolutions.}
We train and evaluate Reloc3r at two image resolutions: widths of 224 and 512 pixels. The results are shown in the tables in Sec.~\ref{sec:relpose} and Sec.~\ref{sec:visloc}. 
Similar to DUSt3R~\cite{wang2024dust3r}, the higher resolution improves accuracy but increases runtime due to processing more tokens. 

\myParagraph{Comparison with asymmetric network architecture.} 
We train an asymmetric version of Reloc3r-512 that uses separate decoders and heads for each branch. As shown in Table~\ref{tab:ablation}, this version performs even worse than the default Reloc3r while requiring more computational resources.

\myParagraph{Comparison with pose prediction with metric scale.} 
We train a version of Reloc3r-512 with metric poses as output. The lower accuracy results reported in Table~\ref{tab:ablation} validate the effectiveness of our default design without metric scales.

\myParagraph{Interesting findings.} 
We visualize the cross-attention responses in some blocks from Reloc3r’s decoder. Interestingly, we find that several layers have developed the ability to match patch correspondences, as illustrated in Figure~\ref{fig:match}.

\myParagraph{Limitations.}
A failure case of Reloc3r occurs when the query image and all the retrieved database images are perfectly collinear, leading to a degeneracy issue that makes the metric scale unsolvable using motion averaging methods.

\vspace{-1pt}

\section{Conclusion}
\label{sec:conclusion}

\vspace{-1pt}

In this paper, we present Reloc3r, a simple yet effective visual localization framework. It consists of an elegantly designed relative pose regression network with a minimalist motion averaging module. Leveraging large-scale training on around eight million image pairs, Reloc3r demonstrates strong generalization capability, high efficiency, and accurate pose estimation performance across multiple datasets, while remaining parsimonious.
We hope it will advance research on data scale and diversity in visual localization. 

\myParagraph{Acknowledgements.} 
This work is supported by 
the Early Career Scheme of the Research Grants Council 
(grant \# 27207224),
the HKU-100 Award, 
a donation from the Musketeers Foundation, and in part by the JC STEM Lab of Autonomous Intelligent Systems funded by The Hong Kong Jockey Club Charities Trust.
This work is also supported in part by the Research Council of Finland for funding the projects (grants No. 353138, No. 362407, and No. 352788). We acknowledge the computational resources provided by the CSC-IT Center for Science, Finland and the HKU-Shanghai ICRC. We thank the reviewers for their valuable feedback. Siyan Dong would like to thank the support from HKU School of Computing and Data Science. 


\appendix
\section*{Supplementary Material}

This appendix provides additional content that cannot be included in the main paper due to page limitations.

\section{Training Details}
Similar to DUSt3R~\cite{wang2024dust3r}, we randomly sample a fixed number of 50K image pairs from each dataset at each training epoch. During training, we augment the image pairs with random color jittering. 
For Reloc3r-512, we begin training directly with images at the maximum resolution of 512 pixels. Within each batch, the image aspect ratios are randomly selected from [4:3, 32:21, 16:9, 2:1, 16:5]. During inference, test image pairs are resized to a width of 512 pixels while maintaining their original aspect ratios.
In contrast, for Reloc3r-224, the image resolution is fixed to 224$\times$224 for both training and inference.

Our symmetric architecture consists of a ViT-Large as the encoder~\cite{dosovitskiy2020image}, a ViT-Base as the decoder, and a regression head. We freeze the ViT encoder and only update the weighs for the decoder and pose regression head during the training. Unlike DUSt3R, which uses both image orders $(I_1, I_2)$ and $(I_2, I_1)$ during training for better generalization, our symmetric design allows us to feed only $(I_1, I_2)$ directly. 
This approach speeds up the training process and reduces memory and storage consumption, which will be discussed in detail in Sec.~\ref{sec:supp_ablation}.

\section{Detailed Ablation Studies}
\label{sec:supp_ablation}

\myParagraph{Symmetric vs. asymmetric networks.} 
DUSt3R~\cite{wang2024dust3r}'s two branches are designed to learn different capabilities. They aim to solve scene reconstruction in a unified coordinate system. For convenience, they choose the first frame's local coordinate system as the unified system. Therefore, the first branch focuses on 3D geometry reconstruction without requiring coordinate transformations, while the second branch handles both geometry reconstruction and coordinate system alignment. In contrast, Reloc3r focuses on learning relative poses, which are inherently symmetric for the two branches. To leverage this property, we adapt DUSt3R's architecture by introducing shared decoder and prediction head, simplifying the model while preserving its effectiveness.

The asymmetric version of Reloc3r follows DUSt3R's design~\cite{wang2024dust3r}, which employs separate decoders and regression heads for the two input images. However, this approach increases the number of learnable parameters and introduces a potential bias based on the image order. To mitigate this bias, DUSt3R incorporates flipped image pairs during training, which adds additional computational overhead. As shown in Table~\ref{tab:ablation} in the main paper, we demonstrate that the asymmetric version performs even worse than the default Reloc3r on the ScanNet1500 dataset~\cite{dai2017scannet,sarlin2020superglue}. 
This underscores the benefits of our fully symmetric architecture, where both branches share decoder and prediction head. Remarkably, our model (with 0.43B parameters) achieves superior accuracy while using approximately 28\% fewer parameters compared to the asymmetric variant.

\myParagraph{Learning relative poses with metric scales?} 
As discussed in the main paper, learning metric scales in relative poses can divert the network's focus from estimating camera orientation and movement direction, potentially hindering generalization across datasets. 
To investigate this, we conduct an ablation study on learning relative poses with metric scales. Following recent works~\cite{winkelbauer2021learning,arnold2022map}, we normalize the translation output as a unit vector and add an additional layer to regress the metric translation scale.
The predicted translation vectors and scales are supervised with the L1 loss. 
We evaluate this version on ScanNet1500~\cite{dai2017scannet,sarlin2020superglue} and Cambridge Landmarks~\cite{kendall2015posenet}. 
The relative pose estimation results are reported in Table~\ref{tab:ablation}. 
Notably, in this setup, the predicted scale factors are irrelevant to the task and we observe a decrease in the accuracy of relative pose estimation compared to our default Reloc3r.
These findings validate the effectiveness of the non-metric design, which allows the network to focus on two critical aspects: camera orientation and movement direction.

The results of absolute pose estimation are presented in Table~\ref{tab:camb_supp}. Methods labeled as metric represent the versions that learn metric camera poses. We observe that the predicted scale estimates lack accuracy, leading to translation errors similar to baseline methods~\cite{winkelbauer2021learning,arnold2022map}.
For further evaluation, we focus solely on translation directions combined with top-2 motion averaging, which produces significantly improved results. This finding validates our approach of estimating metric scales through motion averaging rather than directly learning them with neural networks, highlighting its robustness and effectiveness.

\myParagraph{Rotation representations.} 
We use a continuous 9D-to-SO(3) mapping~\cite{levinson2020analysis} in Reloc3r to avoid the discontinuities found in 3D and 4D representations. In Table~\ref{tab:rot}, we reports an ablation study using different rotation representations. The experiments are trained on ScanNet++~\cite{yeshwanth2023scannet++} and tested on ScanNet1500~\cite{dai2017scannet,sarlin2020superglue}. The results demonstrate the effectiveness of the 9D rotation representation.

\begin{table}[htb!]
\centering
\resizebox{0.37\textwidth}{!}{
\small
\begin{tabular}{l|ccc}
\toprule
Rot. representations & 3D & 4D & 9D (default) \\
\toprule
AUC@20 & 66.81 & 67.87  & \textbf{68.70} \\
\bottomrule
\end{tabular}
}
\caption{
Ablation study for different rotation representations. 
}
\label{tab:rot}
\end{table}

\begin{table}[tbh!]
\centering
\resizebox{0.44\textwidth}{!}{

\small

\begin{tabular}{l|ccc}
\toprule
\multirow{2}{*}{Methods} & \multicolumn{3}{|c}{ScanNet1500} \\
& AUC@5 & AUC@10 & AUC@20 \\
\toprule

No init. (224) & {3.74} & {14.59} & {34.04} \\ 

No init. (512) & {3.98} & {15.58} & {37.02} \\ 

No init. (224 to 512) & {6.76} & {21.96} & {44.38} \\ 

\midrule

DUSt3R-512 (encoder) & 17.83 & 41.08 & 63.05 \\

CroCo v2 (full) & {22.44} & {47.62} & {68.65} \\

MASt3R (full) & 32.62 & 56.28 & 74.32 \\

DUSt3R-512 (full) & \textbf{34.79} & \textbf{58.37} & \textbf{75.56} \\

\bottomrule
\end{tabular}
}
\caption{
Ablations on different network weight initializations. 
}
\label{tab:init}
\end{table}

\myParagraph{Study on the importance of network weight initialization.}
The proposed Reloc3r builds on the recent foundation model DUSt3R~\cite{wang2024dust3r}, leveraging its pre-trained weights for initialization. 
Here, we explore different approaches for network weights initialization: using pre-trained weights from other models, and random initialization. 

Table~\ref{tab:init} presents the test results for these initialization methods. 
Training from MASt3R~\cite{leroy2024grounding} and CroCo~\cite{weinzaepfel2023croco} results in worse pose accuracy. Similarly, when only initializing the encoder part from DUSt3R and training the decoder from scratch, the performance also degrades.
Without pre-trained weights as initialization, we observe a significant drop in performance, a phenomenon similarly observed in DUSt3R trained without CroCo initialization. 
Interestingly, even in the random initialized version, we still can observe meaningful interactions in the cross-attention layers. These layers demonstrate functionality akin to feature matching, despite the absence of ground-truth correspondences for supervision. 
Additional analysis of this behavior is provided in the following Sec.~\ref{sec:supp_analyses}.

\section{More Analyses}
\label{sec:supp_analyses}

\begin{figure}[!tbh]
\centering
\includegraphics[width=0.98\linewidth]{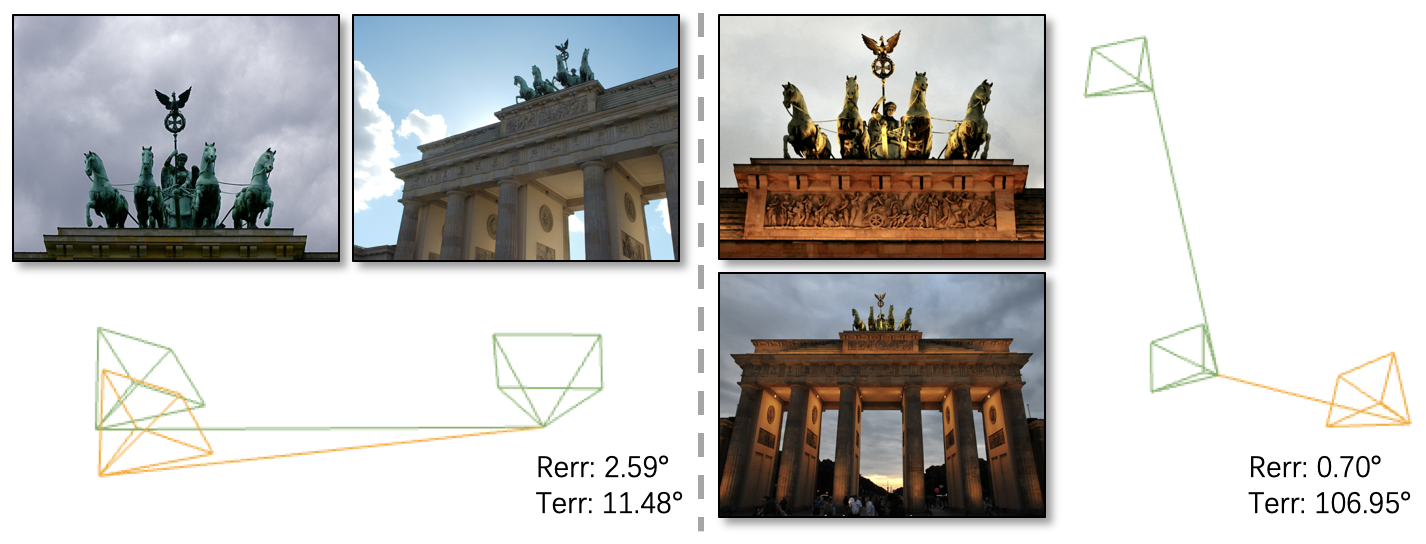}
\caption{
Our pose regression network encounters failure cases when significant changes in focal length occur. As shown in the figure, there are 3$\times$ to 4$\times$ zoom in / out effects. While rotation estimates remain largely unaffected, translation becomes noticeably inaccurate. This issue is similar to the scale-distance ambiguity problem in two-view geometry.
}
\label{fig:fail_intrin}
\end{figure}

\begin{figure}[!tbh]
\centering
\includegraphics[width=0.98\linewidth]{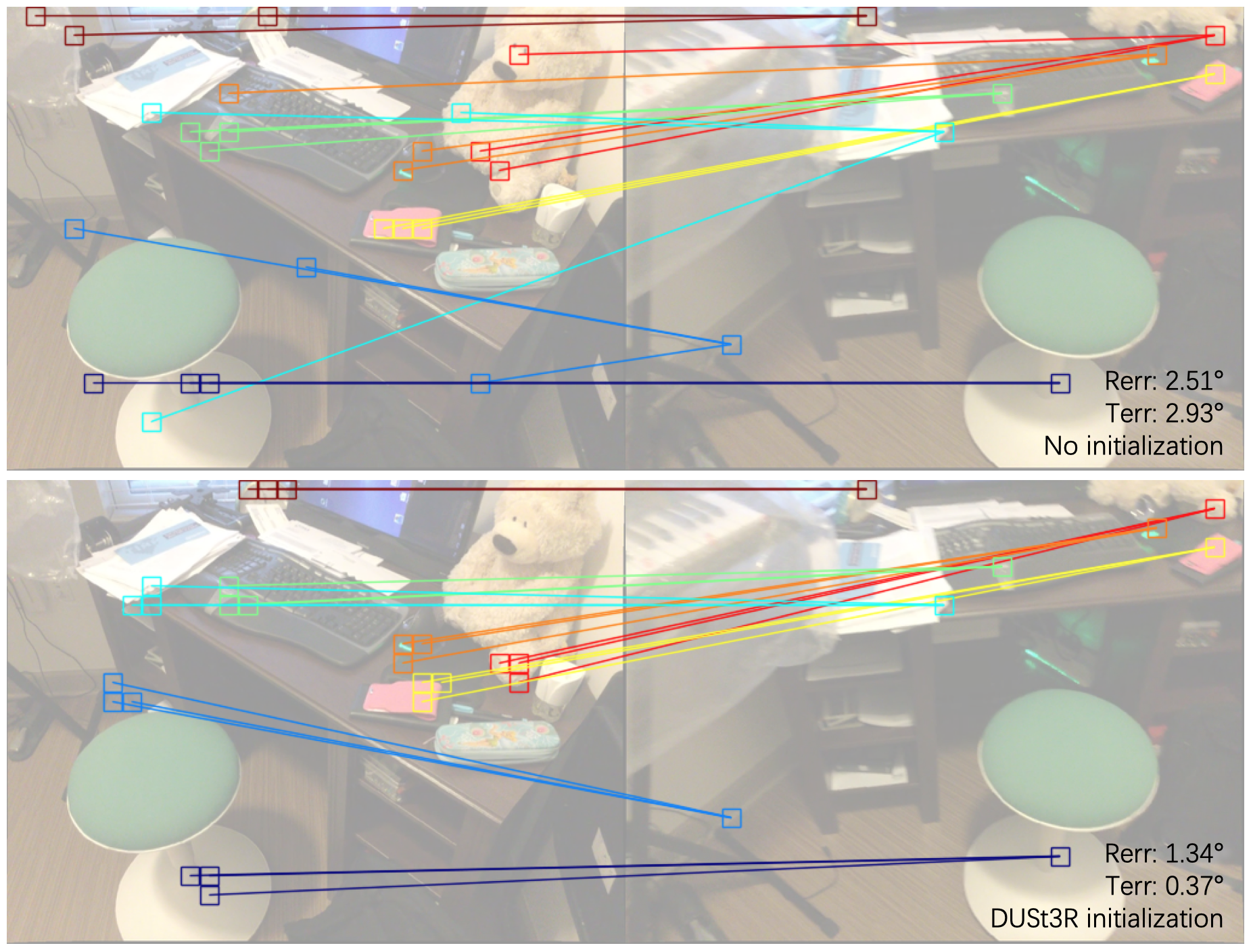}
\caption{
Visualization of top-3 cross-attention responses on the ScanNet1500 dataset~\cite{dai2017scannet,sarlin2020superglue}. 
The top row displays results from Reloc3r trained without pretraining, while the bottom row shows the default Reloc3r trained with DUSt3R initialization. 
}
\label{fig:match_sn}
\end{figure}
\begin{figure}[!tbh]
\centering
\includegraphics[width=0.98\linewidth]{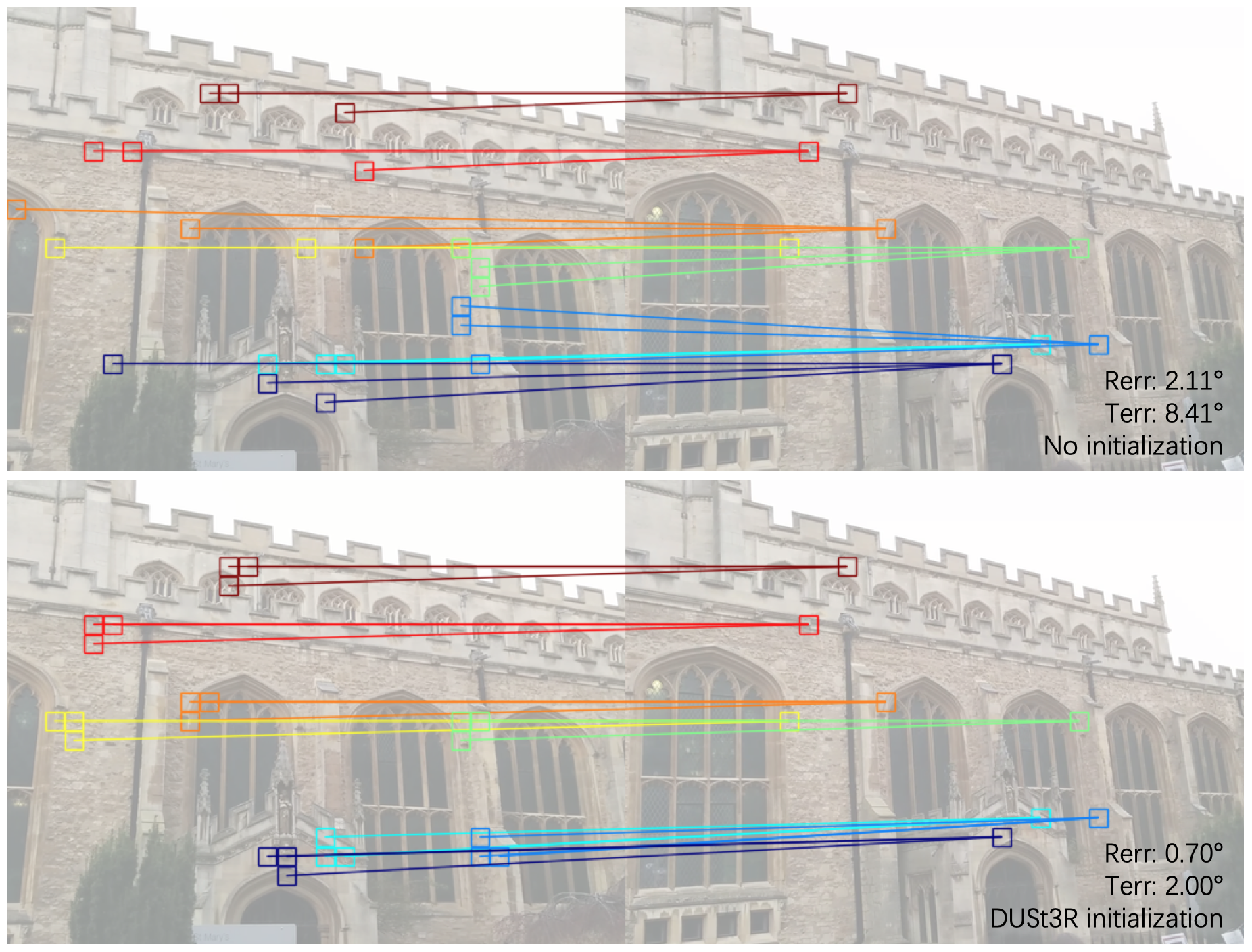}
\caption{
Visualization of top-3 cross-attention responses on the Cambridge Landmarks~\cite{kendall2015posenet}.
The top row displays results from Reloc3r trained without pretraining, while the bottom row shows the default Reloc3r trained with DUSt3R initialization. 
}
\label{fig:match_cl}
\end{figure}

\begin{table*}[!th]
\centering
\resizebox{1.0\textwidth}{!}{

\small

\begin{tabular}{c|l|ccccc|c|c|c}
\toprule
& \multirow{1}{*}{Methods} & \multirow{1}{*}{GreatCourt} & \multirow{1}{*}{KingsCollege} & \multirow{1}{*}{OldHospital} & \multirow{1}{*}{ShopFacade} & \multirow{1}{*}{StMarysChurch} & \multirow{1}{*}{Average (4)} & \multirow{1}{*}{Average} & Inference time \\

\toprule

\multirow{3}{*}{\rotatebox{90}{FM}} 

& HLoc (SP+SG)~\cite{sarlin2019coarse,detone2018superpoint,sarlin2020superglue} 
& \textbf{0.10 / 0.05} 
& \textbf{0.07 / 0.10}
& \textbf{0.13 / 0.23}
& \textbf{0.03 / 0.14}
& \textbf{0.04 / 0.16}
& \textbf{0.07 / 0.16}
& \textbf{0.07 / 0.14}
& {737 ms} \\

& {LazyLoc}~\cite{dong2023lazy} (top-20)
& 0.14 / 0.08
& \textbf{0.07} / 0.13
& 0.20 / 0.37
& 0.04 / 0.15
& 0.06 / 0.18
& 0.09 / 0.21
& 0.10 / 0.18
& {1041 ms} \\

& DUSt3R-512~\cite{wang2024dust3r} (top-20)
& 0.38 / 0.16
& 0.11 / 0.20
& 0.17 / 0.33
& 0.06 / 0.26
& 0.07 / 0.24
& 0.10 / 0.26
& 0.16 / 0.24
& $>$3000 ms \\

\midrule

\multirow{3}{*}{\rotatebox{90}{SCR}} 

& DSAC* (RGB+3D)~\cite{brachmann2021visual}
& 0.49 / 0.3
& \textbf{0.15 / 0.3}
& \textbf{0.21 / 0.4}
& \textbf{0.05 / 0.3}
& \textbf{0.13 / 0.4}
& \textbf{0.14 / 0.4}
& 0.21 / \textbf{0.3}
& - \\

& DSAC* (RGB)~\cite{brachmann2021visual}
& \textbf{0.34 / 0.2}
& 0.18 / \textbf{0.3}
& \textbf{0.21 / 0.4}
& \textbf{0.05 / 0.3}
& 0.15 / 0.6
& 0.15 / \textbf{0.4}
& \textbf{0.19} / 0.4
& - \\

& ACE~\cite{brachmann2023accelerated}
& 0.43 / \textbf{0.2}
& 0.28 / 0.4
& 0.31 / 0.6
& \textbf{0.05 / 0.3}
& 0.18 / 0.6
& 0.21 / 0.5
& 0.25 / 0.4
& - \\

\midrule

\multirow{10}{*}{\rotatebox{90}{RPR}} 

& {Map-free (Regress)}~\cite{arnold2022map} & 8.40 / 4.56 &
2.44 / 2.54 &
3.73 / 5.23 &
0.97 / 3.17 &
2.91 / 5.10 &
2.51 / 4.01 & 
3.69 / 4.12 &
{11 ms} \\

& {ExReNet (SUNCG)}~\cite{winkelbauer2021learning} & 9.79 / 4.46 & 2.33 / 2.48 & 3.54 / 3.49 & 0.72 / 2.41 & 2.30 / 3.72 & 2.22 / 3.03 & 3.74 / 3.31 & {18 ms} \\

& ImageNet+NCM~\cite{zhou2020learn}$\dagger$ & - & - & - & - & - & 0.83 / 1.36 & - & - \\

& \textbf{Reloc3r-224 top-10} 
& 1.71 / 0.94
& 0.47 / 0.41
& 0.87 / 0.66
& 0.18 / \textbf{0.53}
& 0.41 / 0.73
& 0.48 / 0.58
& 0.73 / 0.65
& 51 ms \\

& \textbf{Reloc3r-512 metric} & 9.18 / 1.20 & 2.77 / 0.60 & 3.79 / 0.96 & 0.95 / 0.92 & 2.98 / 0.99 & 2.62 / 0.87 & 3.93 / 0.93 & {42 ms}\\

& \textbf{Reloc3r-512 metric top-2} & 2.86 / 1.18 & 0.95 / 0.53 & 1.41 / 0.86 & 0.37 / 0.79 & 0.63 / 0.91 & {0.84 / 0.77} & {1.24 / 0.85} & 54 ms \\

& \textbf{Reloc3r-512 top-2} & 2.41 / 0.86 & 0.75 / 0.41 & 1.22 / 0.48 & 0.18 / 0.55 & 0.60 / 0.65 & 0.69 / 0.52 & 1.03 / 0.59 & 54 ms \\

& \textbf{Reloc3r-512 top-5} & 1.26 / \textbf{0.72} & 0.49 / 0.39 & 0.77 / 0.54 & \textbf{0.13} / 0.55 & 0.40 / 0.60 & 0.45 / 0.52 & 0.61 / 0.56 & {122 ms} \\ 

& \textbf{Reloc3r-512 top-10} 
& 1.22 / 0.73
& \textbf{0.42 / {0.36}} 
& {0.62 / {0.55}}  
& \textbf{0.13} / 0.58 
& \textbf{0.34} / 0.58  
& \textbf{0.38} / 0.52 & 0.55 / 0.56 & {235 ms} \\

& \textbf{Reloc3r-512 top-10 robust} & \textbf{0.95 / 0.72} & 0.45 / \textbf{0.36} & \textbf{0.58 / 0.53} & \textbf{0.13 / 0.53} & \textbf{0.34 / 0.54} & \textbf{0.38 / 0.49} & \textbf{0.49 / 0.54} & {235 ms} \\

\bottomrule
\end{tabular}

}
\caption{
{Additional results on the Cambridge Landmarks~\cite{kendall2015posenet}.}
Note that although DUSt3R-512 regresses coordinates, it performs pixel-to-pixel matching with these regressed coordinates for accurate visual localization. 
The inference times of Reloc3r are reported using fp32. 
}
\label{tab:camb_supp}
\end{table*}

\myParagraph{Visualization of cross-attention responses.} 
We are interested in how Reloc3r achieves its performance and aim to understand what the network has learned. To this end, we visualize the cross-attention maps in the decoder blocks and observe an interesting behavior: they resemble patch-wise correspondence matching. Results from two datasets are presented in Figure~\ref{fig:match_sn} and Figure~\ref{fig:match_cl}. For clarity, the query patches in the right-hand figures are manually selected for better visualization.

From random initialization, the network still gains the ability to build correspondences, with only relative poses as supervision. When initialized with DUSt3R's pre-trained weights, the cross-attention responses are more accurate and concentrated. This may stem from dense pixel-wise coordinate supervision. We believe introducing ground-truth correspondence information and supervising the across-attention maps could potentially enhance network performance, or accelerate convergence during training.

\myParagraph{Model sizes.} 
Previous works mainly focus on algorithm design, yet we take a different direction by scaling up the training to develop (to the best of our knowledge) the first foundation model for camera pose regression. As a result, Reloc3r's relative pose regression network contains 0.43B parameters - far larger than existing camera pose regression networks (e.g., Map-free with 22M and Marepo with 10M parameters). Despite its size, it achieves real-time inference on consumer-grade GPUs like NVIDIA 3090/4090. We chose Transformer architectures as our backbone for their proven ability to scale better than Convolutional Neural Networks (CNNs). Our experiments with Map-free (ResUNet) showed that its 22M parameters led to underfitting on our training data. Even after expanding the CNN's Resblocks and feature dimensions (up to 0.1B parameters), the model only memorized the training data. All CNN models we tested performed poorly, achieving AUC@20 $<$5 on the ScanNet1500 datasetet. 
While their rotation accuracy can be reasonable, their translation accuracy is poor. 

\myParagraph{Scale and diversity of training data. }
In Table~\ref{tab:train_ablation}, we show that larger training sets consistently improve pose estimation accuracy. Removing domain-specific data (such as the object-centric Co3Dv2 dataset) has minimal impact on accuracy in other domains. This suggests that diverse data helps with generalization, while domain-specific data improves accuracy within its domain.

\begin{table}[htb!]
\centering
\resizebox{0.49\textwidth}{!}{
\small
\begin{tabular}{l|ccc}
\toprule
AUC@20 on datasets & ScanNet1500 & RE10K & ACID \\ 
\toprule
Reloc3r-512 trained w/ ScanNet++ only  & 68.70 & 58.52 & 51.15 \\ 
Reloc3r-512 trained w/o RE10K \& Co3Dv2  & 75.46 & 84.44 & 67.41 \\ 
Reloc3r-512 trained w/o RE10K & {75.55} & {85.33} & {67.76} \\ 
Reloc3r-512 full training & \textbf{75.56} & \textbf{88.39} & \textbf{70.34} \\ 
\bottomrule
\end{tabular}
}
\caption{
Ablation study on training data. 
}
\label{tab:train_ablation}
\end{table}

\myParagraph{Additional discussion on limitations and future works.}
As discussed in the main paper, a primary limitation of Reloc3r is the degeneracy issue of solving the metric translation with motion averaging when all the images are perfectly collinear.  In such cases, the metric scale becomes unsolvable. Although our experiments show that directly regressing metric poses leads to inferior results, this remains an open direction for future research to explore.

While classical feature-matching methods solve relative poses using the 5-point algorithm~\cite{hartley2003multiple} with ground-truth camera intrinsics, our pose regression network does not explore this intrinsic information. This limitation results in some failure cases similar to the scale-distance ambiguity issue (Figure~\ref{fig:fail_intrin}), making it challenging to predict the movement of the camera center. Future research could explore embedding intrinsic parameters directly into the network or regressing the essential matrix as a potential solution.

\begin{figure*}[!tbh]
\centering
\includegraphics[width=1\linewidth]{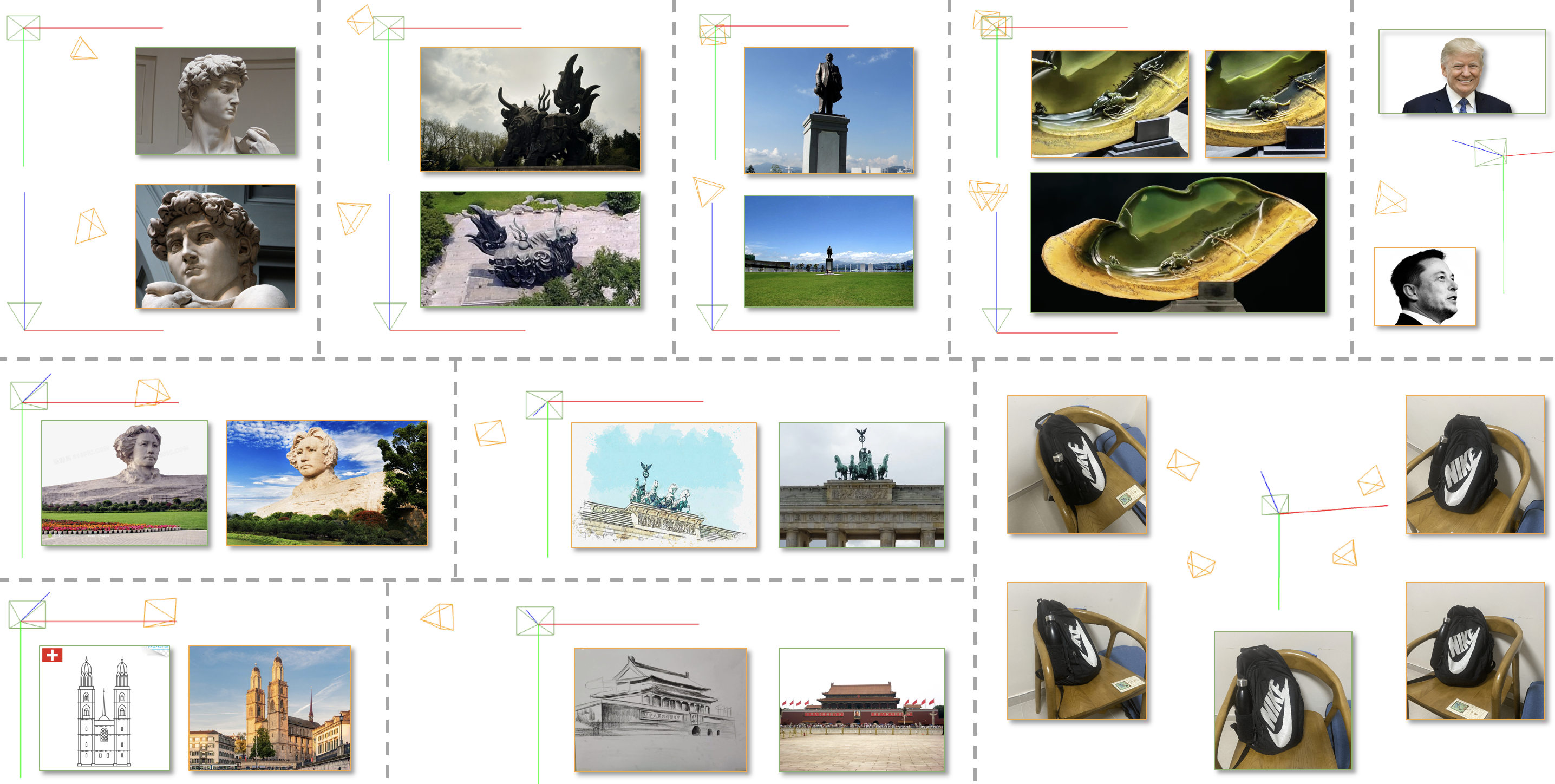}
\caption{
We visualize relative pose estimates using both internet-sourced and self-captured images. For better visualization, we plot the axes of the first view, and the metric scale of the translation vectors is set to 1 meter. 
}
\label{fig:wild_relpose}
\end{figure*}
\begin{figure*}[!tbh]
\centering
\includegraphics[width=1\linewidth]{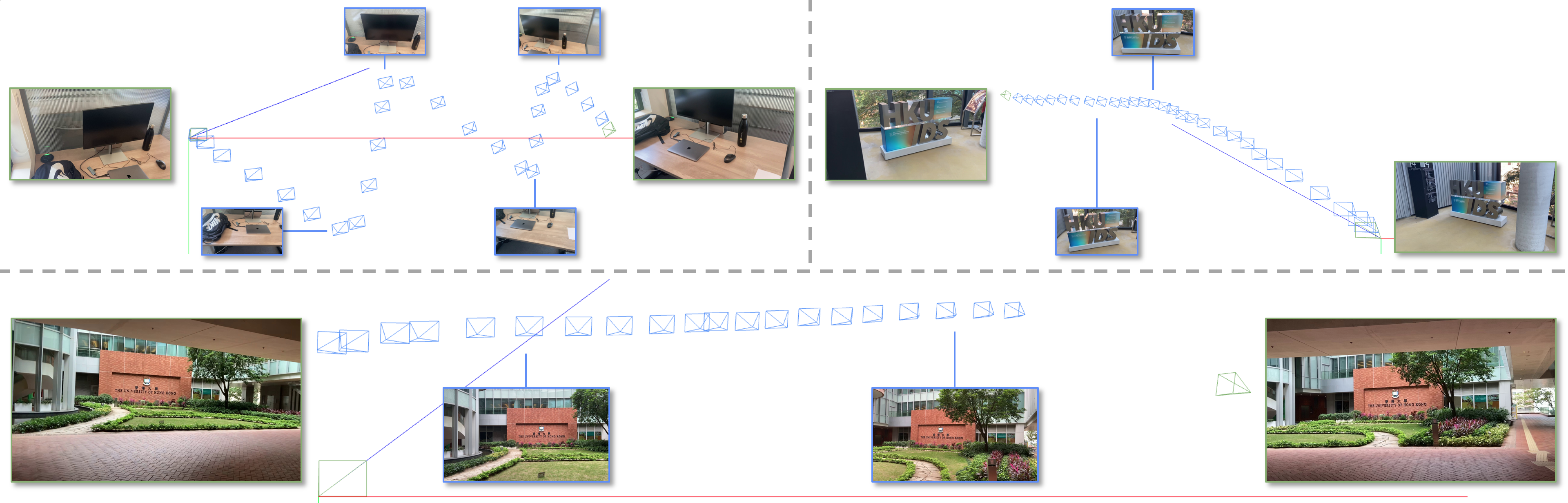}
\caption{
We visualize absolute pose estimates using casually captured videos. For each video, we use two database images whose poses are estimated by our pose regression network. The metric scale of the translation between database images is set to 1 meter. 
}
\label{fig:wild_visloc}
\end{figure*}

\section{Additional Comparisons}

\myParagraph{Relative pose estimation on MegaDepth1500~\cite{li2018megadepth,sun2021loftr}.} 
The results are presented in Table~\ref {tab:mega}. This dataset exhibits significant intrinsic variations between image pairs, which pose a major challenge for pose regression methods and often lead to failures in estimating the translation direction. We also compare our method with matching-based competitors,  where DUSt3R~\cite{wang2024dust3r} and MASt3R~\cite{leroy2024grounding} are evaluated with image resolution 512 $\times$ 512, and the relative poses are obtained from essential matrix estimation in OpenCV~\cite{bradski2000opencv}. While our method achieves reasonable pose accuracy, it still falls short compared to SoTA matching-based approaches. Figure~\ref{fig:fail_intrin} illustrates some failure cases, which are also discussed in Sec.~\ref{sec:supp_analyses}. 

\begin{table}[th!]
\centering
\resizebox{0.48\textwidth}{!}{

\small

\begin{tabular}{c|l|ccc}

\toprule

&\multirow{2}{*}{Methods} & \multicolumn{3}{|c}{MegaDepth1500} \\
& & AUC@5 & AUC@10 & AUC@20 \\

\toprule

\multirow{5}{*}{\rotatebox{90}{Non-PR}} 

& Efficient LoFTR~\cite{wang2024efficient} & 56.4 & 72.2 & 83.5 \\

& ROMA~\cite{edstedt2024roma} & \textbf{62.6} & \textbf{76.7} & \textbf{86.3} \\

& DUSt3R~\cite{wang2024dust3r} & 27.9 & 46.0 & 63.3\\

& MASt3R~\cite{leroy2024grounding} & 42.4 & 61.5 & 76.9 \\

\midrule

\multirow{6}{*}{\rotatebox{90}{PR}} 

& Map-free (Regress-SN)~\cite{arnold2022map} & - & - & $<$10 \\

& Map-free (Regress-MF)~\cite{arnold2022map} & - & - & $<$10 \\

& ExReNet (SN)~\cite{winkelbauer2021learning} & - & - & $<$10 \\

& ExReNet (SUNCG)~\cite{winkelbauer2021learning} & - & - & $<$10 \\

& \textbf{Reloc3r-224} & {39.9} & {59.7} & {75.4} \\
& \textbf{Reloc3r-512} & \textbf{49.6} & \textbf{67.9} & \textbf{81.2} \\

\bottomrule
\end{tabular}
}
\caption{
Relative camera pose evaluation on the MegaDepth1500 dataset~\cite{li2018megadepth,sun2021loftr}.
}
\label{tab:mega}
\end{table}

\myParagraph{Comparison with FAR~\cite{rockwell2024far}.} 
Recent works FAR~\cite{rockwell2024far} and PanoPose~\cite{tu2024panopose} design pose regression networks for wide baseline pairs and panorama images. While FAR performs well on images with few overlaps, it underperforms Reloc3r on popular datasets used in the main paper. Specifically, we tested FAR on ScanNet1500, RE10K, and ACID datasets, achieving AUC@20 of 28.19, 37.67, and 44.98\%, respectively. Since PanoPose has not released its code yet, we look forward to comparing with it in the future. 

\myParagraph{Visual localization with different experimental settings.} 
We conduct these experiments on the Cambridge Landmarks~\cite{kendall2015posenet}. The results are shown in Table~\ref{tab:camb_supp}. 

In our evaluation of metric pose estimation, we compare results with and without motion averaging. Due to the challenge of learning metric scales, using top-2 motion averaging yields significantly better results compared to single pairs. For Reloc3r-512, we test varying numbers of top-$K$ image pairs. While increasing the number of images reduces error, it also leads to longer inference times. We also try to adopt LazyLoc~\cite{dong2023lazy}'s rotation and translation averaging modules as robust estimators. These provide limited improvements across most scenes, with the notable exception of GreatCourt, which features extensive repetitive patterns and similar regions.
Since Reloc3r does not produce matches, it cannot adopt the post-optimization step used in LazyLoc. Like other pose regression-based methods, Reloc3r therefore still underperforms in pose accuracy compared to SoTA feature matching-based methods on large-scale scenes. 
The accuracy of pose regression also can not match with those of scene coordinate regression (SCR) based methods, as SCR methods typically require per-scene training and can take long inference times.

\section{Details for the Compared Methods}
For relative pose estimation on ScanNet1500~\cite{dai2017scannet, sarlin2020superglue}, Re10K~\cite{zhou2018stereo}, and ACID~\cite{liu2021infinite}.
In NoPoSplat's implementation, images are first resized and center-cropped to 256$\times$256, then upscaled to 560$\times$560 at the coarse level, and finally to 864$\times$864 to match ROMA~\cite{edstedt2024roma}'s settings. Our approach, however, maintains original aspect ratios while limiting maximum image resolution to 512px. For DUSt3R~\cite{wang2024dust3r} and MASt3R~\cite{leroy2024grounding}, different from NoPoSplat that uses the input resolution of 512$\times$256, we set it to 512$\times$512. On MegaDepth1500~\cite{li2018megadepth,sun2021loftr}, evaluation resolutions also vary across methods, following their original settings. For example, Efficient LoFTR~\cite{wang2024efficient} is evaluated with an image resolution of 1200$\times$1200, RoMA uses 560$\times$560, while our method employs a resolution of 512px.
For the PR-based competitors, We report the pose regression versions of Map-free~\cite{arnold2022map} trained on ScanNet~\cite{dai2017scannet} and their Map-free dataset. Similarly, we evaluate two versions of ExReNet trained on ScanNet and SUNCG~\cite{song2017semantic}. 

For multi-view pose estimation on CO3Dv2~\cite{reizenstein2021common}, we randomly sample 10 images from each test sequence to form 45 pairs, yielding 76,905 total pairs for evaluation. For RayReg~\cite{zhang2024cameras} and RayDiffusion~\cite{zhang2024cameras}, we report the results based on the 8-view setup described in the paper, as we could not produce reasonable results with 10 views. 

For absolute metric pose estimation on 7 Scenes~\cite{shotton2013scene} and Cambridge~\cite{kendall2015posenet}, the results mainly come from the original publication of each paper, except Map-free and ExReNet. We evaluate two versions of Map-free: regression and hybrid with matching. For 7 Scenes, we use checkpoints trained on ScanNet, while for the Cambridge dataset, we use checkpoints trained on the Map-free dataset to maintain consistency between indoor and outdoor settings. 
For ExReNet, we also evaluate their two versions on both 7 Scenes and Cambridge datasets.

For the remaining methods not covered above, we cite results directly from their original publications.

\section{In-The-Wild Camera Pose Estimations}
We test Reloc3r with ``in-the-wild'' images and videos collected from the internet and captured by ourselves.

The results for relative pose estimation are shown in Figure~\ref{fig:wild_relpose}. Thanks to large-scale training, we find that Reloc3r generalizes well across diverse viewpoint changes and can infer relative poses between paintings, sketches, and real images. Surprisingly, it achieves reasonable results even when processing the faces of different people. 

The results for visual localization are shown in Figure~\ref{fig:wild_visloc}. For each video, we use two images as a database to localize query images in the video. The database poses are estimated by our pose regression network. Note that when the database and query images are collinear, the metric scale cannot be reliably recovered due to the degeneracy issue.

{
\small
\bibliographystyle{ieeenat_fullname}
\bibliography{main}
}


\end{document}